\documentclass{article}

\usepackage[final]{neurips_2020}

\usepackage[utf8]{inputenc} 
\usepackage[T1]{fontenc}    
\usepackage{url}            
\usepackage{booktabs}       
\usepackage{amsfonts}       
\usepackage{nicefrac}       
\usepackage{microtype}      
\usepackage{dsfont}

\usepackage{xcolor}
\definecolor{cgreen}{rgb}{0.2,0.6,0.2}
\definecolor{darkred}{rgb}{0.4,0.0,0.0}
\definecolor{darkgreen}{rgb}{0.0,0.4,0.0}
\definecolor{darkblue}{rgb}{0.0,0.0,0.4}
\usepackage{hyperref}
\hypersetup{colorlinks, citecolor=darkblue, urlcolor=darkblue}

\usepackage{graphicx}
\usepackage{subcaption}
\usepackage{wrapfig}

\graphicspath{{figures/classes/}{figures/illustrations/}}

\usepackage{multirow}
\usepackage{makecell}

\usepackage[ruled]{algorithm2e}
\usepackage{listings,newtxtt}

\definecolor{deepblue}{rgb}{0,0,0.6}
\definecolor{deepred}{rgb}{0.6,0,0}
\definecolor{deepgreen}{rgb}{0,0.5,0}

\lstdefinestyle{lststyle}{
language=Python,
basicstyle=\ttfamily\small,
commentstyle=\color{deepred},
otherkeywords={self},             
keywordstyle=\color{deepgreen},
emph={__init__,forward,inverse,VAE,Augment,UniformDequantization,Flow,Normal,CouplingBijection,Reverse},          
emphstyle=\color{deepblue},    
stringstyle=\color{deepred},
frame=tb,                         
showstringspaces=false            %
}

\usepackage{amsmath}
\usepackage{amssymb}

\newcommand{\vk}{\boldsymbol{k}}

\newcommand{\vu}{\boldsymbol{u}}

\newcommand{\vx}{\boldsymbol{x}}

\newcommand{\vz}{\boldsymbol{z}}

\newcommand{\vI}{\boldsymbol{I}}
\newcommand{\vJ}{\boldsymbol{J}}

\newcommand{\vW}{\boldsymbol{W}}

\newcommand{\vpi}{\boldsymbol{\pi}}

\newcommand{\cB}{\mathcal{B}}

\newcommand{\cE}{\mathcal{E}}

\newcommand{\cI}{\mathcal{I}}

\newcommand{\cL}{\mathcal{L}}

\newcommand{\cN}{\mathcal{N}}

\newcommand{\cS}{\mathcal{S}}

\newcommand{\cV}{\mathcal{V}}

\newcommand{\cX}{\mathcal{X}}
\newcommand{\cY}{\mathcal{Y}}
\newcommand{\cZ}{\mathcal{Z}}

\newcommand{\bE}{\mathbb{E}}

\newcommand{\bI}{\mathbb{I}}

\newcommand{\KL}{\mathbb{D}_{\mathsf{KL}}}
\DeclareMathOperator*{\argmax}{arg\,max}

\DeclareMathOperator*{\sign}{sign}
\DeclareMathOperator*{\sort}{sort}
\DeclareMathOperator*{\argsort}{argsort}
\DeclareMathOperator*{\Cat}{Cat}
\DeclareMathOperator*{\Bern}{Bern}
\DeclareMathOperator*{\Unif}{Unif}

\newcommand{\RR}{\mathds{R}}

\newcommand{\grad}{{\nabla}}

\newcommand{\ie}{{i.e.}\xspace}

\newcommand{\cf}{{cf.}\xspace}

\newtheorem{example}{Example}

\usepackage{cleveref}
\title{SurVAE Flows: Surjections to\\Bridge the Gap between VAEs and Flows}

\author{%
  Didrik Nielsen$^1$, $\,$Priyank Jaini$^2$, $\,$Emiel Hoogeboom$^2$, $\,$Ole Winther$^1$, $\,$Max Welling$^2$ \\
  Technical University of Denmark$^1$, $\,$UvA-Bosch Delta Lab, University of Amsterdam$^2$ \\
  \texttt{didni@dtu.dk, p.jaini@uva.nl, e.hoogeboom@uva.nl} \\
  \texttt{olwi@dtu.dk, m.welling@uva.nl}
}

\begin{document}

\maketitle

\begin{abstract}

Normalizing flows and variational autoencoders are powerful generative models that can represent complicated density functions. However, they both impose constraints on the models: Normalizing flows use bijective transformations to model densities whereas VAEs learn stochastic transformations that are non-invertible and thus typically do not provide tractable estimates of the marginal likelihood.
In this paper, we introduce SurVAE Flows: A modular framework of composable transformations that encompasses VAEs and normalizing flows. SurVAE Flows bridge the gap between normalizing flows and VAEs with \emph{surjective transformations}, wherein the transformations are deterministic in one direction -- thereby allowing exact likelihood computation, and stochastic in the reverse direction -- hence providing a lower bound on the corresponding likelihood. 
We show that several recently proposed methods, including dequantization and augmented normalizing flows, can be expressed as SurVAE Flows.
Finally, we introduce common operations such as the \emph{max value}, the \emph{absolute value}, \emph{sorting} and \emph{stochastic permutation} as composable layers in SurVAE Flows.

\end{abstract}

\section{Introduction}
\label{sec:introduction}

Normalizing flows \citep{TabakVE10, TabakTurner13, rezende2015} provide a powerful \emph{modular} and \emph{composable} framework for representing expressive probability densities via differentiable bijections (with a differentiable inverse). These composable bijective transformations accord significant advantages due to their ability to be implemented using a modular software framework with a general interface consisting of three important components: (i) a forward transform, (ii) an inverse transform, and (iii) a log-likelihood contribution through the Jacobian determinant. Thus, significant advances have been made in recent years to develop novel flow modules that are easily invertible, expressive and computationally cheap \citep{dinh2014, dinh2017, kingma2016, papamakarios2017, huang2018, jaini2019tails, jaini2019, kingma2018, hoogeboom2019conv, hoogeboom2020convexp, durkan2019, berg2018}.

However, the bijective nature of the transformations used for building normalizing flows limit their ability to alter dimensionality, model discrete data and distributions with discrete structure or disconnected components. Specialized solutions have been developed to address these limitations independently.  \citet{uria2013, ho2019} use dequantization to model discrete distributions using continuous densities, while \citet{tran2019,hoogeboom2019} propose a discrete analog of normalizing flows. \citet{cornish2019localised} use an augmented space to model an infinite mixtures of normalizing flows to address the problem of disconnected components whereas \citet{huang2020, chen2020} use a similar idea of augmentation of the observation space to model expressive distributions. VAEs \citep{kingma2013, rezende2014}, on the other hand, have no such limitations, but only provide lower bound estimates of the tractable estimates for the exact marginal density. These shortcomings motivate the question: \emph{Is it possible to have composable and modular architectures that are expressive, model discrete and disconnected structure, and allow altering dimensions with exact likelihood evaluation?}

\begin{figure}[t]
    \centering
    \begin{subfigure}[t]{0.245\textwidth}
        \centering
        \resizebox{\textwidth}{!}{
\begingroup%
  \makeatletter%
  \providecommand\color[2][]{%
    \errmessage{(Inkscape) Color is used for the text in Inkscape, but the package 'color.sty' is not loaded}%
    \renewcommand\color[2][]{}%
  }%
  \providecommand\transparent[1]{%
    \errmessage{(Inkscape) Transparency is used (non-zero) for the text in Inkscape, but the package 'transparent.sty' is not loaded}%
    \renewcommand\transparent[1]{}%
  }%
  \providecommand\rotatebox[2]{#2}%
  \newcommand*\fsize{\dimexpr\f@size pt\relax}%
  \newcommand*\lineheight[1]{\fontsize{\fsize}{#1\fsize}\selectfont}%
  \ifx\svgwidth\undefined%
    \setlength{\unitlength}{69.19634421bp}%
    \ifx\svgscale\undefined%
      \relax%
    \else%
      \setlength{\unitlength}{\unitlength * \real{\svgscale}}%
    \fi%
  \else%
    \setlength{\unitlength}{\svgwidth}%
  \fi%
  \global\let\svgwidth\undefined%
  \global\let\svgscale\undefined%
  \makeatother%
  \begin{picture}(1,0.91964142)%
    \lineheight{1}%
    \setlength\tabcolsep{0pt}%
    \put(0,0){\includegraphics[width=\unitlength,page=1]{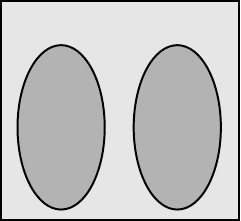}}%
    \put(0.20137517,0.7798154){\color[rgb]{0,0,0}\makebox(0,0)[lt]{\lineheight{1.25}\smash{\begin{tabular}[t]{l}$\cZ$\end{tabular}}}}%
    \put(0.68198678,0.77981572){\color[rgb]{0,0,0}\makebox(0,0)[lt]{\lineheight{1.25}\smash{\begin{tabular}[t]{l}$\cX$\end{tabular}}}}%
    \put(0,0){\includegraphics[width=\unitlength,page=2]{classes_bijection.pdf}}%
  \end{picture}%
\endgroup%

        }
        \caption{Bijective}
    \end{subfigure}
    \begin{subfigure}[t]{0.245\textwidth}
        \centering
        \resizebox{\textwidth}{!}{
\begingroup%
  \makeatletter%
  \providecommand\color[2][]{%
    \errmessage{(Inkscape) Color is used for the text in Inkscape, but the package 'color.sty' is not loaded}%
    \renewcommand\color[2][]{}%
  }%
  \providecommand\transparent[1]{%
    \errmessage{(Inkscape) Transparency is used (non-zero) for the text in Inkscape, but the package 'transparent.sty' is not loaded}%
    \renewcommand\transparent[1]{}%
  }%
  \providecommand\rotatebox[2]{#2}%
  \newcommand*\fsize{\dimexpr\f@size pt\relax}%
  \newcommand*\lineheight[1]{\fontsize{\fsize}{#1\fsize}\selectfont}%
  \ifx\svgwidth\undefined%
    \setlength{\unitlength}{69.19634421bp}%
    \ifx\svgscale\undefined%
      \relax%
    \else%
      \setlength{\unitlength}{\unitlength * \real{\svgscale}}%
    \fi%
  \else%
    \setlength{\unitlength}{\svgwidth}%
  \fi%
  \global\let\svgwidth\undefined%
  \global\let\svgscale\undefined%
  \makeatother%
  \begin{picture}(1,0.91964142)%
    \lineheight{1}%
    \setlength\tabcolsep{0pt}%
    \put(0,0){\includegraphics[width=\unitlength,page=1]{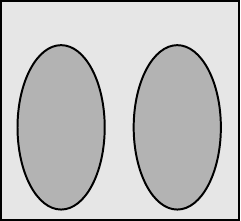}}%
    \put(0.20137517,0.7798154){\color[rgb]{0,0,0}\makebox(0,0)[lt]{\lineheight{1.25}\smash{\begin{tabular}[t]{l}$\cZ$\end{tabular}}}}%
    \put(0.68198678,0.77981572){\color[rgb]{0,0,0}\makebox(0,0)[lt]{\lineheight{1.25}\smash{\begin{tabular}[t]{l}$\cX$\end{tabular}}}}%
    \put(0,0){\includegraphics[width=\unitlength,page=2]{classes_surjection_gen.pdf}}%
  \end{picture}%
\endgroup%

        }
        \caption{Surjective (Gen.)}
    \end{subfigure}
    \begin{subfigure}[t]{0.245\textwidth}
        \centering
        \resizebox{\textwidth}{!}{
\begingroup%
  \makeatletter%
  \providecommand\color[2][]{%
    \errmessage{(Inkscape) Color is used for the text in Inkscape, but the package 'color.sty' is not loaded}%
    \renewcommand\color[2][]{}%
  }%
  \providecommand\transparent[1]{%
    \errmessage{(Inkscape) Transparency is used (non-zero) for the text in Inkscape, but the package 'transparent.sty' is not loaded}%
    \renewcommand\transparent[1]{}%
  }%
  \providecommand\rotatebox[2]{#2}%
  \newcommand*\fsize{\dimexpr\f@size pt\relax}%
  \newcommand*\lineheight[1]{\fontsize{\fsize}{#1\fsize}\selectfont}%
  \ifx\svgwidth\undefined%
    \setlength{\unitlength}{69.19634421bp}%
    \ifx\svgscale\undefined%
      \relax%
    \else%
      \setlength{\unitlength}{\unitlength * \real{\svgscale}}%
    \fi%
  \else%
    \setlength{\unitlength}{\svgwidth}%
  \fi%
  \global\let\svgwidth\undefined%
  \global\let\svgscale\undefined%
  \makeatother%
  \begin{picture}(1,0.91964142)%
    \lineheight{1}%
    \setlength\tabcolsep{0pt}%
    \put(0,0){\includegraphics[width=\unitlength,page=1]{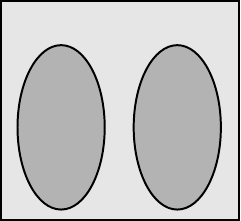}}%
    \put(0.20137517,0.7798154){\color[rgb]{0,0,0}\makebox(0,0)[lt]{\lineheight{1.25}\smash{\begin{tabular}[t]{l}$\cZ$\end{tabular}}}}%
    \put(0.68198678,0.77981572){\color[rgb]{0,0,0}\makebox(0,0)[lt]{\lineheight{1.25}\smash{\begin{tabular}[t]{l}$\cX$\end{tabular}}}}%
    \put(0,0){\includegraphics[width=\unitlength,page=2]{classes_surjection_inf.pdf}}%
  \end{picture}%
\endgroup%

        }
        \caption{Surjective (Inf.)}
    \end{subfigure}
    \begin{subfigure}[t]{0.245\textwidth}
        \centering
        \resizebox{\textwidth}{!}{
\begingroup%
  \makeatletter%
  \providecommand\color[2][]{%
    \errmessage{(Inkscape) Color is used for the text in Inkscape, but the package 'color.sty' is not loaded}%
    \renewcommand\color[2][]{}%
  }%
  \providecommand\transparent[1]{%
    \errmessage{(Inkscape) Transparency is used (non-zero) for the text in Inkscape, but the package 'transparent.sty' is not loaded}%
    \renewcommand\transparent[1]{}%
  }%
  \providecommand\rotatebox[2]{#2}%
  \newcommand*\fsize{\dimexpr\f@size pt\relax}%
  \newcommand*\lineheight[1]{\fontsize{\fsize}{#1\fsize}\selectfont}%
  \ifx\svgwidth\undefined%
    \setlength{\unitlength}{69.19634421bp}%
    \ifx\svgscale\undefined%
      \relax%
    \else%
      \setlength{\unitlength}{\unitlength * \real{\svgscale}}%
    \fi%
  \else%
    \setlength{\unitlength}{\svgwidth}%
  \fi%
  \global\let\svgwidth\undefined%
  \global\let\svgscale\undefined%
  \makeatother%
  \begin{picture}(1,0.91964142)%
    \lineheight{1}%
    \setlength\tabcolsep{0pt}%
    \put(0,0){\includegraphics[width=\unitlength,page=1]{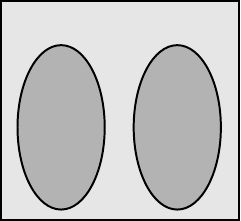}}%
    \put(0.20137517,0.7798154){\color[rgb]{0,0,0}\makebox(0,0)[lt]{\lineheight{1.25}\smash{\begin{tabular}[t]{l}$\cZ$\end{tabular}}}}%
    \put(0.68198678,0.77981572){\color[rgb]{0,0,0}\makebox(0,0)[lt]{\lineheight{1.25}\smash{\begin{tabular}[t]{l}$\cX$\end{tabular}}}}%
    \put(0,0){\includegraphics[width=\unitlength,page=2]{classes_stochastic.pdf}}%
  \end{picture}%
\endgroup%

        }
        \caption{Stochastic}
    \end{subfigure}
    \caption{
    Classes of SurVAE transformations $\cZ \rightarrow \cX$ and their inverses $\cX \rightarrow \cZ$. Solid lines indicate deterministic transformations, while dashed lines indicate stochastic transformations.
    }
    \label{fig:classes}
\end{figure}

In this paper, we answer this affirmatively by introducing SurVAE Flows that use surjections to provide a unified, composable, and modular framework for probabilistic modeling. We introduce our unifying framework in \S\ref{sec:premlim} by identifying the components necessary to build composable architectures with modular software implementation for probabilistic modeling. We then introduce surjections for probabilistic modeling in \S\ref{sec:survae} and show that these transformations lie at the interface between VAEs (stochastics maps) and normalizing flows (bijective maps). We unify these transformations (bijections, surjections, and stochastic transformations) in a composable and modular framework that we call SurVAE Flows. Subsequently, in \S\ref{subsec:novel}, we propose novel SurVAE Flow layers like \emph{max value} used for max pooling layers, \emph{absolute value} for modelling symmetries in the data, and \emph{sorting} and \emph{stochastic permutations} that can be used for modelling exchangeable data and order statistics. Finally, in \S\ref{subsec:connec} we connect SurVAE Flows to several aforementioned specialised models by expressing them using SurVAE Flow layers which can now be implemented easily using our modular implementation. We demonstrate the efficacy of SurVAE Flows with experiments on synthetic datasets, point cloud data, and images. Code to implement SurVAE Flows and reproduce results is publicly available\footnote{The code is available at \url{https://github.com/didriknielsen/survae_flows}}.

\section{Preliminaries and Setup}
\label{sec:premlim}

In this section, we set up our main problem, provide key notations and definitions, and formulate a unifying framework for using different kinds of transformations to model distributions.

Let $\vx \in \cX$ and $\vz \in \cZ$ be two variables with distributions $p(\vx)$ and $p(\vz)$. We call a deterministic mapping $f: \cZ \to \cX$ \emph{bijective} if it is both \emph{surjective} and \emph{injective}. 
A mapping is surjective if $\forall \vx \in \cX$, $\exists~ \vz \in \cZ$ such that $\vx = f(\vz)$.
A mapping is injective if $\forall \vz_1, \vz_2 \in \cZ$, $f(\vz_1) = f(\vz_2) \implies \vz_1 = \vz_2$.
If the mapping is not deterministic, we refer to it as a stochastic mapping, and denote it as $\vx \sim p(\vx|\vz)$.

Normalizing flows \citep{TabakVE10,TabakTurner13, rezende2015} make use of bijective transformations $f$ to transform a simple base density $p(\vz)$ to a more expressive density $p(\vx)$, making using the change-of-variables formula $p(\vx) = p(\vz)|\det\nabla_{\vx}f^{-1}(\vx)|$. 
VAEs \citep{kingma2013, rezende2014}, on the other hand, define a probabilistic graphical model where each observed variable $\vx$ has an associated latent variable $\vz$ with the generative process as $\vz \sim p(\vz), ~\vx \sim p(\vx|\vz)$, where $p(\vx|\vz)$ may be viewed as a stochastic transformation. VAEs use variational inference with an amortized variational distribution $q(\vz | \vx)$ to approximate the intractable posterior $p(\vz | \vx)$ which facilitates computation of a lower bound of $p(\vx)$ known as the evidence lower bound (ELBO) \ie, $\cL := \bE_{q(\vz|\vx)}[\log p(\vx|\vz)] - \KL [q(\vz|\vx) \| p(\vz)]$.

In the following, we introduce a framework to connect flows and VAEs\footnote{We note that \citet{wu2020} also considered stochastic maps in flows using MCMC transition kernels.} by showing that bijective and stochastic transformations are \emph{composable} and require three important components for use in probabilistic modeling: 
(i) a forward transformation, $f:\cZ \to \cX$ with an associated conditional probability $p(\vx|\vz)$, (ii) an inverse transformation, $f^{-1} : \cX \to \cZ$ with an associated distribution $q(\vz|\vx)$, and (iii) a \emph{likelihood contribution} term used for log-likelihood computation. 

\begin{table}[ht]
\caption{Composable building blocks of SurVAE Flows.}
\label{tab:flow_layers}
\centering
\begin{tabular}{l|c|c|c|c}
\toprule
\multirow{2}{*}{\textbf{Transformation}} & \textbf{Forward} & \textbf{Inverse} & \textbf{Likelihood Contribution} & \textbf{Bound Gap} \\
 & $\vx \leftarrow \vz$ & $\vz \leftarrow \vx$ & $\cV(\vx, \vz)$ & $\cE(\vx, \vz)$ \\
\midrule
Bijective & $\vx = f(\vz)$ & $\vz = f^{-1}(\vx)$ & $\log |\det \grad_{\vx} \vz|$ & 0 \\
\midrule
Stochastic  & $\vx \sim p(\vx | \vz)$ & $\vz \sim q(\vz | \vx)$ & $\log \frac{p(\vx | \vz)}{q(\vz | \vx)}$ & $\log \frac{q(\vz | \vx)}{p(\vz|\vx)}$ \\
\midrule
Surjective (Gen.) & $\vx = f(\vz)$ & $\vz \sim q(\vz|\vx)$ & $\log \frac{p(\vx | \vz)}{q(\vz | \vx)} \footnotesize{\,\text{as}\,\,} \scriptsize{\begin{aligned}
&p(\vx | \vz) \rightarrow \\
&\delta(\vx - f(\vz))
\end{aligned}}$ & $\log \frac{q(\vz | \vx)}{p(\vz|\vx)}$\\
\midrule
Surjective (Inf.) & $\vx \sim p(\vx | \vz)$ & $\vz = f^{-1}(\vx)$ & $\log \frac{p(\vx | \vz)}{q(\vz | \vx)} \footnotesize{\,\text{as}\,\,} \scriptsize{\begin{aligned}
&q(\vz | \vx) \rightarrow \\
&\delta(\vz - f^{-1}(\vx))
\end{aligned}}$ & 0 \\
\bottomrule
\end{tabular}
\end{table}

\textbf{Forward Transformation:} For a stochastic transformation, the forward transformation is the conditional distribution
$p(\vx | \vz)$. For a bijective transformation, on the other hand, the forward transformation is deterministic and therefore, $p(\vx|\vz) = \delta\big(\vx - f(\vz)\big)$ or simply $\vx = f(\vz)$.

\textbf{Inverse Transformation:} For a bijective transformation, the inverse is also deterministic and given by $\vz = f^{-1}(\vx)$. For a stochastic transformation, the inverse is also stochastic and is defined by Bayes theorem $p(\vz | \vx) = p(\vx|\vz)p(\vz)/p(\vx)$. Computing $p(\vz|\vx)$ is typically intractable and we thus resort to a variational approximation $q(\vz|\vx)$. 

\textbf{Likelihood Contribution:} For bijections, the density $p(\vx)$ can be computed from $p(\vz)$ and the mapping $f$ using the change-of-variables formula as:
\begin{equation}
\label{eq:cov}
    \log p(\vx) = \log p(\vz) + \log |\det \vJ |, \qquad \vz = f^{-1}(\vx)
\end{equation}
where $|\det \vJ| = |\det \grad_{\vx} f^{-1}(\vx)|$ is the absolute value of the determinant of the Jacobian matrix of $f^{-1}$ which defines the likelihood contribution term for a bijective transformation $f$. For stochastic transformations, we can rewrite the marginal density $p(\vx)$ as:
\begin{equation}
    \label{eq:st}
    \log p(\vx) = \underbrace{\bE_{q(\vz|\vx)}[\log p(\vx|\vz)] - \KL [q(\vz|\vx) \| p(\vz)]}_{\mathrm{ELBO}} + \underbrace{\KL [q(\vz|\vx) \| p(\vz|\vx)]}_{\mathrm{Gap\, in\, Lower\, Bound}}
\end{equation}
\begin{wrapfigure}{r}{0.35\textwidth}
\vspace{-4mm}
\begin{algorithm}[H]
\label{algo:loglik}
 \KwData{$\vx$, $p(\vz)$ \& $\{f_t\}_{t=1}^T$}
 \KwResult{$\cL(\vx)$}
 \For{$t$ in $\mathsf{range}(T),$}{
 \uIf{$f_t$ is bijective}{
  $\vz = f_t^{-1}(\vx)$ \;
  $\cV_t = \log \left| \det \frac{\partial \vz}{\partial \vx}\right|$ \;
 }
 \uElseIf{$f_t$ is stochastic}{
  $\vz \sim q_t(\vz|\vx)$ \;
  $\cV_t = \log \frac{p_t(\vx|\vz)}{q_t(\vz|\vx)}$ \;
 }
 $\vx = \vz$ \;
 }
 \Return{$\log p(\vz) + \sum_{t=1}^T \cV_t$}
 \caption{$\mathsf{log-likelihood}(\vx)$}
\end{algorithm}
\end{wrapfigure}
The ELBO $\cL$ in Eq.~\ref{eq:st} can then be evaluated using a single Monte Carlo sample: $\cL \approx \log p(\vz) + \log \frac{p(\vx|\vz)}{q(\vz | \vx)},~ \vz \sim q(\vz|\vx)$.
Therefore, the likelihood contribution term for a stochastic transformation is defined as $\log \frac{p(\vx|\vz)}{q(\vz | \vx)}$. Furthermore, we show in App.~\ref{app:vae_limit} that Eq.~\ref{eq:st} allows us to recover the change-of-variables formula given in Eq.~\ref{eq:cov} by using Dirac delta functions, thereby drawing a precise connection between VAEs and normalizing flows. 
Crucially, Eq.~\ref{eq:st} helps us to reveal a unified modular framework to model a density $p(\vx)$ under any transformation by restating it as:
\begin{equation}
\label{eq:comp}
    \log p(\vx) \simeq
    \log p(\vz) + \cV(\vx, \vz) + \cE(\vx, \vz), \quad \vz \sim q(\vz|\vx) 
\end{equation}
where $\cV(\vx, \vz)$ and $\cE(\vx,\vz)$ are the \emph{likelihood contribution} and \emph{bound looseness} terms, respectively. 
The likelihood contribution is $\cV(\vx, \vz)=\log |\det \vJ|$ for bijections and $\log \frac{p(\vx|\vz)}{q(\vz | \vx)}$ for stochastic transformations. For bijections, likelihood evaluation is deterministic and  \emph{exact} with $\cE(\vx, \vz) = 0$, while for stochastic maps it is stochastic and unbiased with a bound looseness of $\cE(\vx, \vz) = \log \frac{q(\vz | \vx)}{p(\vz | \vx)}$. This is summarized in Table \ref{tab:flow_layers}.
The first term in Eq.~\ref{eq:comp}, $\log p(\vz)$, reveals the compositional nature of the transformations, since it can be modeled by further transformations. While the compositional structure has been used widely for bijective transformations, Eq.~\ref{eq:comp} demonstrates its viability for stochastic maps as well. We demonstrate this unified compositional structure in Alg.~\ref{algo:loglik}.

\newpage
\section{SurVAE Flows}
\label{sec:survae}

As explained in \Cref{sec:premlim}, bijective and stochastic transformations provide a modular framework for constructing expressive generative models. However, they both impose constraints on the model: bijective transformations are deterministic and allow exact likelihood computation, but they are required to preserve dimensionality. On the other hand, stochastic transformations are capable of altering the dimensionality of the random variables but only provide a stochastic lower bound estimate of the likelihood. \emph{Is it possible to have composable transformations that can alter dimensionality and allow exact likelihood evaluation?} In this section, we answer this question affirmatively by introducing surjective transformations as SurVAE Flows that bridge the gap between bijective and stochastic transformations.

In the following, we will define composable deterministic transformations that are surjective and non-injective. For brevity, we will refer to them as \emph{surjections} or \emph{surjective transformations}. 
Note that for surjections, multiple inputs can map to a single output, resulting in a \emph{loss of information} since the input is not guaranteed to be recovered through inversion. Similar to bijective and stochastic transformations, the three important components of composable surjective transformations are:

\textbf{Forward Transformation:} Like bijections, surjective transformations have a deterministic forward transformation $p(\vx | \vz) = \delta \big(\vx - f(\vz)\big)$ or $\vx = f(\vz)$.

\textbf{Inverse Transformation:} In contrast with bijections, surjections $f:\cZ \to \cX$ are not invertible since multiple inputs can map to the same output. 
However, they have \emph{right inverses}, i.e.~functions $g: \cX \to \cZ$ such that $f \circ g (\vx) = \vx$, but not necessarily  $g \circ f (\vz) = \vz$. 
We will use a stochastic right inverse $q(\vz|\vx)$ which can be thought of as passing $\vx$ through a random right inverse $g$. Importantly, $q(\vz|\vx)$ only has support over the preimage of $\vx$, i.e. the set of $\vz$ that map to $\vx$,~$\cB(\vx) = \{\vz | \vx=f(\vz)\}$. 

So far, we have described what we will term \emph{generative surjections}, i.e.~transformations that are surjective in the generative direction $\cZ \rightarrow \cX$. We will refer to a transformation which is surjective in the inference direction $\cX \rightarrow \cZ$ as an \emph{inference surjection}. These are illustrated in Fig.\ref{fig:classes}. Generative surjections have stochastic inverse transformations $q(\vz|\vx)$, while inference surjections have stochastic forward $p(\vx|\vz)$ transformations.

\textbf{Likelihood Contribution:} For continuous surjections, the likelihood contribution term is:
\begin{align*}
    \label{eq:lc_sur}
     \bE_{q(\vz | \vx)}\left[\log\frac{p(\vx|\vz)}{q(\vz|\vx)}\right], \quad \text{ as } 
     \begin{cases}
     \quad p(\vx | \vz) \to \delta\big(\vx - f(\vz)\big), \quad\, \text{for gen. surjections.} \\
     \quad q(\vz | \vx) \to \delta\big(\vz - f^{-1}(\vx)\big), \, \text{for inf. surjections.}
     \end{cases}
\end{align*}
While generative surjections generally give rise to stochastic estimates of the likelihood contribution and introduce lower bound likelihood estimates, inference surjections  allow \emph{exact likelihood computation} (see App. \ref{app:bound_looseness}). Before proceeding further, we give a few examples to better understand the construction of a surjective transformation for probabilistic modeling.

\begin{example}[Tensor slicing] Let $f$ be a tensor slicing surjection that takes input $\vz = (\vz_1, \vz_2) \in \RR^{d_z}$ and returns a subset of the elements, i.e.~$\vx = f(\vz) = \vz_1$. To develop this operation as a SurVAE layer, we first specify the stochastic forward and inverse transformations as:
\begin{align*}
    p(\vx|\vz) = \cN(\vx|\vz_1, \sigma^2\vI), \quad \text{ and } \quad 
    q(\vz|\vx) = \cN(\vz_1|\vx, \sigma^2\vI) q(\vz_2|\vx)
\end{align*}
We next compute the likelihood contribution term in the limit that $ p(\vx | \vz) \to \delta\big(\vx - f(\vz)\big)$. Here, this corresponds to $\sigma \to 0$. Thus,
\begin{equation*}
    \cV(\vx, \vz) = \lim_{\sigma^2 \rightarrow 0} \bE_{q(\vz|\vx)}\left[ \log \frac{p(\vx|\vz)}{q(\vz|\vx)} \right] = \bE_{q(\vz_2|\vx)}\left[ - \log  q(\vz_2|\vx) \right],
\end{equation*}
which corresponds to the entropy of $q(\vz_2|\vx)$ that is used to infer the sliced elements $\vz_2$. We illustrate the slicing surjection for both the generative and inference directions in Fig.~\ref{fig:surjections}.
\label{ex:slicing}
\end{example}

\begin{example}[Rounding] Let $f$ be a rounding surjection that takes an input $\vz \in \RR^{d_{\vz}}$ and returns the rounded $\vx := \lfloor \vz \rfloor$. The forward transformation is a discrete surjection $P(\vx|\vz) = \bI(\vz \in \cB(\vx))$, for $\cB(\vx) = \{\vx+\vu|\vu\in[0,1)^d\}$. The inverse transformation $q(\vz|\vx)$ is stochastic with support in $\cB(\vx)$. 
Inserting this in the likelihood contribution term and simplifying, we find
\begin{equation*}
    \cV(\vx, \vz) = \bE_{q(\vz|\vx)}\left[ - \log  q(\vz|\vx) \right].
\end{equation*}
This generative rounding surjection gives rise to dequantization \citep{uria2013, ho2019} which is a method commonly used to train continuous flows on discrete data such as images.
\label{ex:rounding}
\end{example}

The preceding discussion shows that surjective transformations can be composed to construct expressive transformations for density modelling.
We call a single surjective transformation a SurVAE layer and a composition of bijective, surjective, and/or stochastic transformations a SurVAE Flow. The unified framework of SurVAE Flows allows us to construct generative models learned using the likelihood (or its lower bound) of the data, utilizing Eq.~\ref{eq:comp}, and \Cref{tab:flow_layers}.

\begin{table}[t]
\caption{Summary of selected inference surjection layers. See App. \ref{app:layers} for more SurVAE layers.}
\label{tab:selected_surjections}
\footnotesize
\centering
\begin{tabular}{l|c|c|c}
\toprule
\textbf{Surjection} & \textbf{Forward} & \textbf{Inverse} & $\cV(\vx, \vz)$ \\ \midrule
\multirow{2}{*}{Abs} & $s \sim \Bern(\pi(z))$ & $s = \sign x$ & \multirow{2}{*}{$\log p(s|z)$} \\
 &  $x = s \cdot z, ~~ s \in \{-1, 1\}$ & $z = |x|$ &  \\ \midrule
\multirow{2}{*}{Max} & $k \sim \Cat(\vpi(z))$ & $k=\argmax \vx$ & \multirow{2}{*}{$\log p(k|z) + \log p(\vx_{-k} | z, k)$} \\
 &  $x_k = z, \vx_{-k}  \sim p(\vx_{-k} | z, k)$ & $z = \max \vx$ &  \\ \midrule
\multirow{2}{*}{Sort} & $\cI \sim \Cat(\vpi(\vz))$ & $\cI = \argsort \vx$ & \multirow{2}{*}{$\log p(\cI|\vz)$} \\
 & $\vx = \vz_{\cI}$ & $\vz = \sort \vx$ &  \\ 
\bottomrule
\end{tabular}
\end{table}
\normalsize

\subsection{Novel SurVAE Layers}
\label{subsec:novel}

\begin{wrapfigure}[23]{r}{0.3\linewidth}
    \vspace{-6mm}
    \begin{subfigure}{.29\textwidth}
    \includegraphics[width=\textwidth]{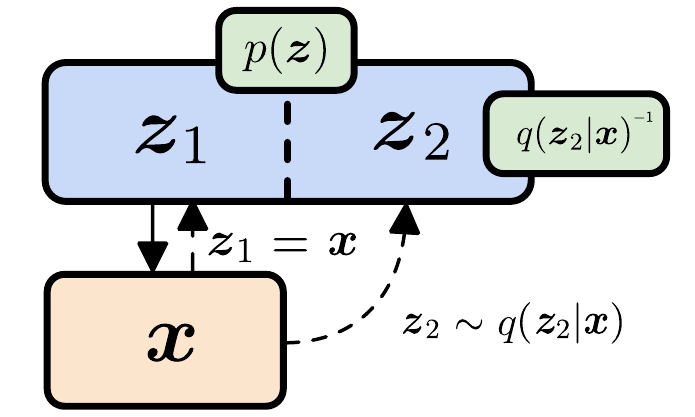}
    \caption{Gen. slicing}
    \end{subfigure}
    \begin{subfigure}{.29\textwidth}
    \includegraphics[width=\textwidth]{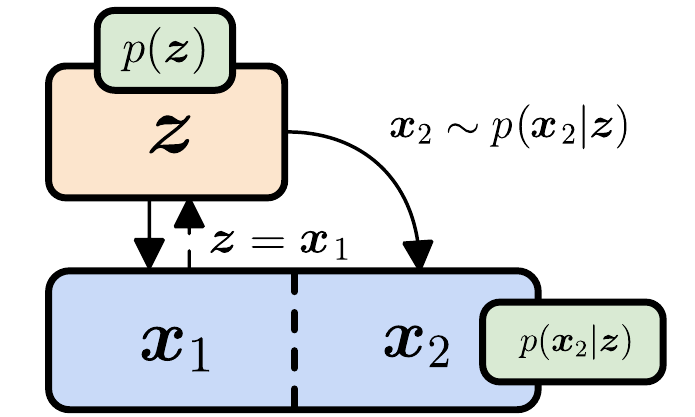}
    \caption{Inf. slicing}
    \end{subfigure}
    \begin{subfigure}{.29\textwidth}
    \includegraphics[width=\textwidth]{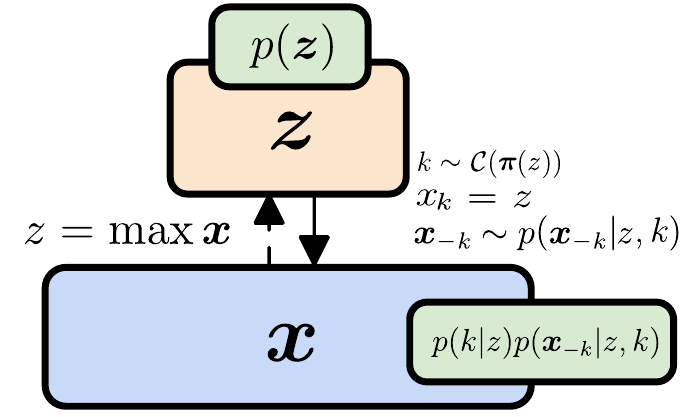}
    \caption{Inf. max}
    \end{subfigure}
    \caption{Surjections.}
    \label{fig:surjections}
\end{wrapfigure}

We developed the tensor slicing and rounding surjections in Examples \ref{ex:slicing} and \ref{ex:rounding}. In this section, we introduce additional novel SurVAE layers including the \emph{absolute value}, the \emph{maximum value} and \emph{sorting} as surjective layers and \emph{stochastic permutation} as a stochastic layer. We provide a summary of these in \Cref{tab:selected_surjections}. Due to space constraints, we defer the derivations and details on each of these surjections to \Cref{app:abs}-\ref{app:permute} along with detailed tables on generative and inference surjections in \Cref{tab:generative_surjections} and \ref{tab:inference_surjections}. 

\textbf{Abs Surjection} (App.~\ref{app:abs}). The abs surjection returns the the magnitude of its input, $z = |x|$. As a SurVAE layer, we can represent the inference surjection with the forward and inverse transformations as:
\footnotesize
\begin{align*}
    p(x|z) &= \sum_{s\in\{-1,1\}} p(\vx|z,s)p(s|z) = \sum_{s\in\{-1,1\}}\delta(x-sz)p(s|z), \\
    q(z|x) &= \sum_{s\in\{-1,1\}} q(z|\vx,s)p(s|x) = \sum_{s\in\{-1,1\}}\delta(z-sx)\delta_{s,\sign(x)}
\end{align*}
\normalsize
where $q(z|x)$ is deterministic corresponding to $z=|x|$. The forward transformation $p(x|z)$ involves the following steps: (i) sample the sign $s$, conditioned on $z$, and (ii) apply the sign to $z$ to obtain $x=sz$. 
Abs surjections are useful for modelling data with symmetries which we demonstrate in our experiments. 

\textbf{Max Surjection} (App.~\ref{app:max}, Fig.~\ref{fig:surjections}). The max operator returns the largest element of an input vector, $z = \max \vx$. We can represent this transformation as
\footnotesize
\begin{align*}
    p(\vx|z) &= \sum_{k=1}^K p(\vx|z,k)p(k|z) = \sum_{k=1}^K \delta(x_k-z)p(\vx_{-k}|z,k)p(k|z), \\
    q(z|\vx) &= \sum_{k=1}^K q(z|\vx,k)q(k|x) = \sum_{k=1}^K \delta(z-x_k)\delta_{k,\argmax(\vx)},
\end{align*}
\normalsize
where $q(z|x)$ is deterministic and corresponds to $z=\max \vx$. While the inverse is deterministic
The stochastic forward proceeds by (i) sampling an index $k$ and setting $x_k = z$, and (ii) imputing the remaining values $\vx_{-k}$ of $\vx$ such that they are all smaller than $x_k$. Max surjections are useful in implementing the \emph{max pooling} layer commonly used in convolutional architectures for downsampling. In our experiments, we demonstrate the use of max surjections for probabilistic modelling of images.

\textbf{Sort Surjection} (App.~\ref{app:sort}). Sorting, $\vz = \mathrm{sort}(\vx)$ returns a vector in sorted order. It is a surjective (and non-injective) transformation since the original order of the vector is lost in the operation even though the dimensions remain the same. 
Sort surjections are useful in modelling naturally sorted data, learning order statistics, and learning an exchangeable model using flows.

\textbf{Stochastic Permutation} (App.~\ref{app:permute}). A stochastic permutation transforms the input vector by shuffling the elements randomly. The inverse pass for a permutation is the same as the forward pass with the likelihood contribution term equal to zero, $\cV=0$. 
Stochastic permutations helps to enforce permutation invariance \ie any flow can be made permutation invariant by adding a final permutation SurVAE layer. 
In our experiments, we compare sorting surjections and stochastic permutations to enforce permutation invariance for modelling exchangeable data.

\textbf{Stochastic Inverse Parameterization}. For surjections, the stochastic inverses have to be defined so that they form a distribution over the possible right-inverses.
Different right-inverse distributions do not have to align over subsets of the space. Consequently, for more sophisticated choices of right-inverses, the log-likelihood may be discontinuous across boundaries of these subsets.
Since these points have measure zero, this does not influence the validity of the log-likelihood. However, it may impede optimization using gradient-based methods.
In our experiments, we did not encounter any specific issues, but for a more thorough discussion see \citep{dinh2019rad}.

\begin{table}[b]
\centering
\caption{SurVAE Flows as a unifying framework.}
\label{tab:unifying}
\footnotesize
\begin{tabular}{c|l}
\toprule
\textbf{Model} & \textbf{SurVAE Flow architecture} \\ 
\toprule
\makecell{Probabilistic PCA \citep{tipping1999probabilistic} \\
          VAE \citep{kingma2013, rezende2014} \\
          Diffusion Models \citep{dickstein2015, ho2020}} &$\cZ \xrightarrow{\mathsf{stochastic}} \cX$\\
\midrule
Dequantization \citep{uria2013, ho2019} & $\cZ \xrightarrow{\mathsf{round}} \cX$\\
\midrule
ANFs, VFlow \citep{huang2020, chen2020} & $\cX \xrightarrow{\mathsf{augment}} \cX \times \cE \xrightarrow{\mathsf{bijection}} \cZ$\\
\midrule
Multi-scale Architectures \citep{dinh2017} & $\cX \xrightarrow{\mathsf{bijection}} \cY \times \cE \xrightarrow{\mathsf{slice}} \cY \xrightarrow{\mathsf{bijection}} \cZ$\\
\midrule
\makecell{CIFs, Discretely Indexed Flows, DeepGMMs \\
    \citep{cornish2019localised, duan2019transport, van2015locally}} & $\cX \xrightarrow{\mathsf{augment}} \cX \times \cE \xrightarrow{\mathsf{bijection}} \cZ \times \cE \xrightarrow{\mathsf{slice}} \cZ$\\
\midrule
RAD Flows \citep{dinh2019rad} & $\cX \xrightarrow{\mathsf{partition}} \cX_{\cE} \times \cE \xrightarrow{\mathsf{bijection}} \cZ \times \cE \xrightarrow{\mathsf{slice}} \cZ$\\
\bottomrule
\end{tabular}
\vspace{-1.5em}
\end{table}
\normalsize

\subsection{Connection to Previous Work}
\label{subsec:connec}

The results above provide a unified framework based on SurVAE Flows for estimating probability densities. We now connect this general approach to several recent works on generative modelling.

The differentiable and bijective nature of transformations used in normalizing flows limit their ability to alter dimensionality, model discrete data, and distributions with disconnected components.
Specialized solutions have been proposed in recent years to address these individually. We now show that these works can be expressed using SurVAE Flow layers, as summarized in Table \ref{tab:unifying}.

\subsubsection{Using Stochastic Transformations}

As discussed in Sec. \ref{sec:premlim}, VAEs \citep{kingma2013, rezende2014} may be formulated as composable stochastic transformations. Probabilistic PCA \cite{tipping1999probabilistic} can be considered a simple special case of VAEs wherein the forward transformation is linear-Gaussian, \ie $p(\vx|\vz) = \cN(\vx|\vW\vz, \sigma^2\vI)$. Due to the linear-Gaussian formulation, the posterior $p(\vz|\vx)$ is tractable and we can thus perform exact stochastic inversion for this model. Diffusion models \citep{dickstein2015, ho2020} are another class of models closely related to VAEs. For diffusion models, the inverse $q(\vz|\vx)$ implements a diffusion step, while the forward transformation $p(\vx|\vz)$ learns to reverse the diffusion process. \citet{wu2020} propose an extended flow framework consisting of bijective and stochastic transformations using MCMC transition kernels. Their method utilizes the same computation as in the general formulation in Algorithm \ref{algo:loglik}, but does not consider surjective maps or an explicit connection to VAEs. Their work shows that MCMC kernels may also be implemented as stochastic transformations in SurVAE Flows. 

\subsubsection{Using Surjective Transformations}

Dequantization \citep{uria2013, ho2019} is used for training continuous flow models on ordinal discrete data such as images and audio. Dequantization fits into the SurVAE Flow framework as a composable generative rounding surjection (\cf Example \ref{ex:rounding}) and thus simplifies implementation. When the inverse $q(\vz|\vx)$ is a standard uniform distributon, \emph{uniform dequantization} is obtained, while a more flexible flow-based distribution $q(\vz|\vx)$ yields \emph{variational dequantization} \citep{ho2019}. 

\begin{figure}[h]
    \centering

\begin{subfigure}[b]{0.31\textwidth}
    \centering
    \includegraphics[width=\textwidth]{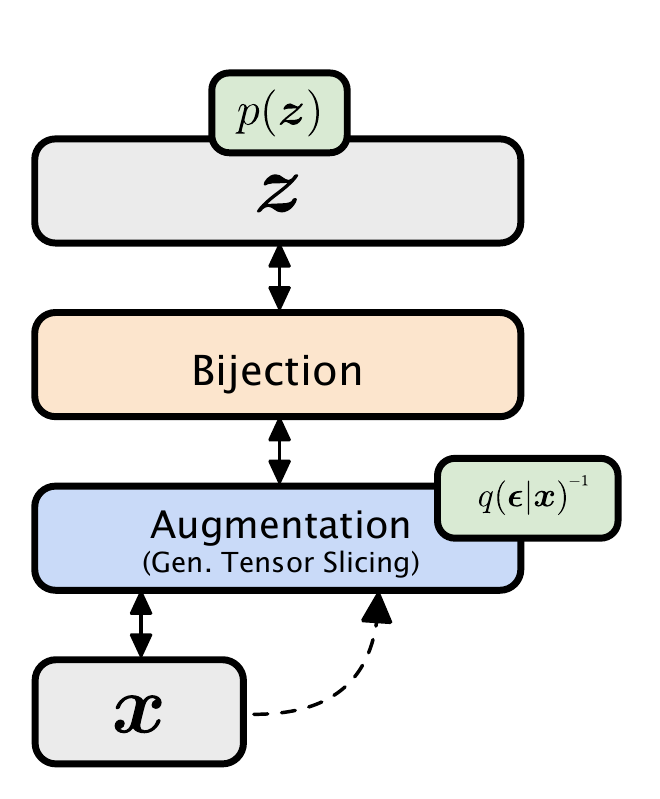}
    \caption{Augmented Flow.}
    \label{fig:arch_aug}
\end{subfigure}
\qquad
\begin{subfigure}[b]{0.31\textwidth}
    \centering
    \includegraphics[width=\textwidth]{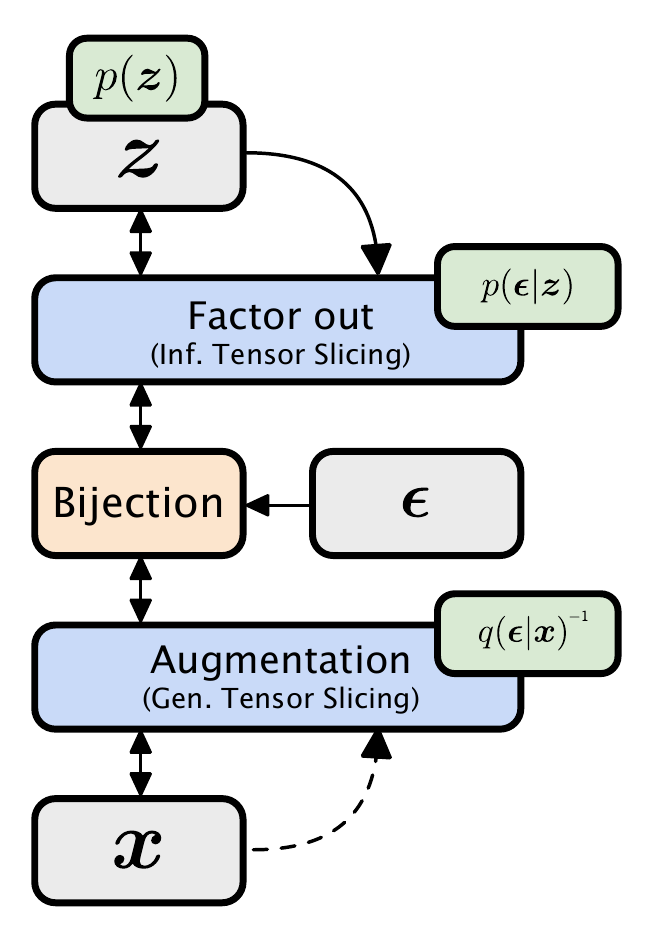}
    \caption{Infinite Mixture of Flows.}
    \label{fig:arch_cmix}
\end{subfigure}

    \caption{Flow architectures making use of tensor slicing.}
    \label{fig:architectures}
\end{figure}

VFlow \citep{chen2020} and ANFs \citep{huang2020} aim to build expressive generative models by augmenting the data space and jointly learning a normalizing flow for the augmented data space as well as the distribution of augmented dimensions. This strategy was also adopted by \citet{dupont2019} 
for continuous-time flows. VFlow and ANFs can be obtained as SurVAE Flows by composing a bijection with a generative tensor slicing surjection (\cf \Cref{ex:slicing} and \Cref{fig:arch_aug}).
The reverse transformation, i.e.~inference slicing, results in the \emph{multi-scale architecture} of \citet{dinh2017}.

CIFs \citep{cornish2019localised} use an indexed family of bijective transformations $g(\cdot ; \varepsilon) : \cZ \to \cX$ where $\cZ = \cX \subseteq \RR^{d}$, and $\varepsilon \in \cE \subseteq \RR^{d_{\varepsilon}}$ with the generative process as: $\vz \sim p(\vz), ~ \epsilon \sim p(\epsilon | \vz)$ and $\vx = g(\vz ; \varepsilon)$ and requires specifying $p(\vz)$ and $p(\varepsilon|\vz)$. CIFs are akin to modeling densities using an infinite mixture of normalizing flows since $g$ is a surjection from an augmented space $\cZ \times \cE$ to the data space $\cX$. Consequently, CIFs can be expressed as a SurVAE flow using a augment surjection composed with a bijection and tensor slicing (\cf \Cref{fig:arch_cmix}). Similarly, \citet{duan2019transport} used a \emph{finite} mixture of normalizing flows to model densities by using a discrete index set $\cE = \{1, 2, 3, \cdots, K\}$ with bijections $g(\cdot ; \varepsilon)$.
Deep Gaussian mixture models \citep{van2015locally} form special case wherein the bijections $g(\cdot ; \varepsilon)$ are linear transformations. 

RAD flows \citep{dinh2019rad} are also ``similar'' to CIFs but it partitions the data space into finite disjoint subsets $\{\cX_i\}_{i=1}^K \subseteq \cX$ and defines bijections $g_i : \cZ \to \cX_i, \forall ~i \in \{1,2,...,K\}$ with the generative process as $\vz \sim p(\vz), i \sim p(i|\vz)$ and $\vx = g_{i}(\vz)$. Interestingly, RAD can be seen to implement a class of inference surjections that rely on partitioning of the data space. The partitioning is learned during training, thus allowing learning of expressive inference surjections. However, careful parameterization is required for stable gradient-based training.
We note that the abs and sort inference surjections introduced earlier may be expressed using a static (non-learned) partitioning of the data space $\cX$ and thus have close ties to RAD. However, RAD does not express generative surjections or more general inference surjections that do not rely on partitioning, such as dimensional changes.

Finally, we note that apart from providing a general method for modelling densities, SurVAE Flows provide a modular framework for easy implementation of the methods described here. We discuss these important software perspectives using code snippets in App.~\ref{app:software}.

\section{Experiments}

\newcommand{\fw}{0.31\linewidth}
\begin{wrapfigure}[32]{r}{0.47\linewidth}
\vspace{-10mm}
\centering
\begin{subfigure}[b]{\fw}
    \caption*{Data}
    \vspace{-2mm}
    \includegraphics[width=\textwidth]{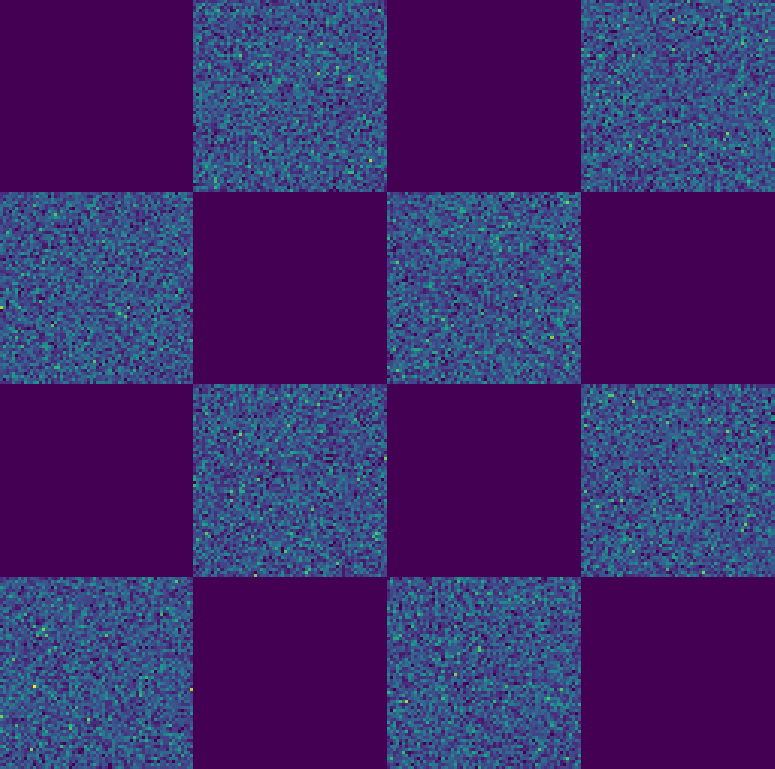}
\end{subfigure}
\begin{subfigure}[b]{\fw}
    \caption*{Flow}
    \vspace{-2mm}
    \includegraphics[width=\textwidth]{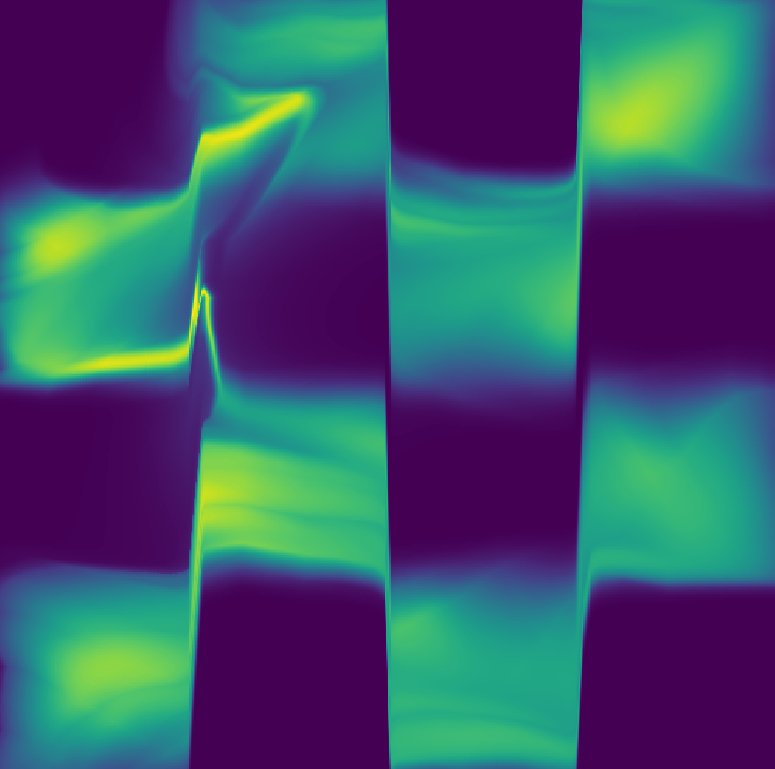}
\end{subfigure}
\begin{subfigure}[b]{\fw}
    \caption*{AbsFlow (ours)}
    \vspace{-2mm}
    \includegraphics[width=\textwidth]{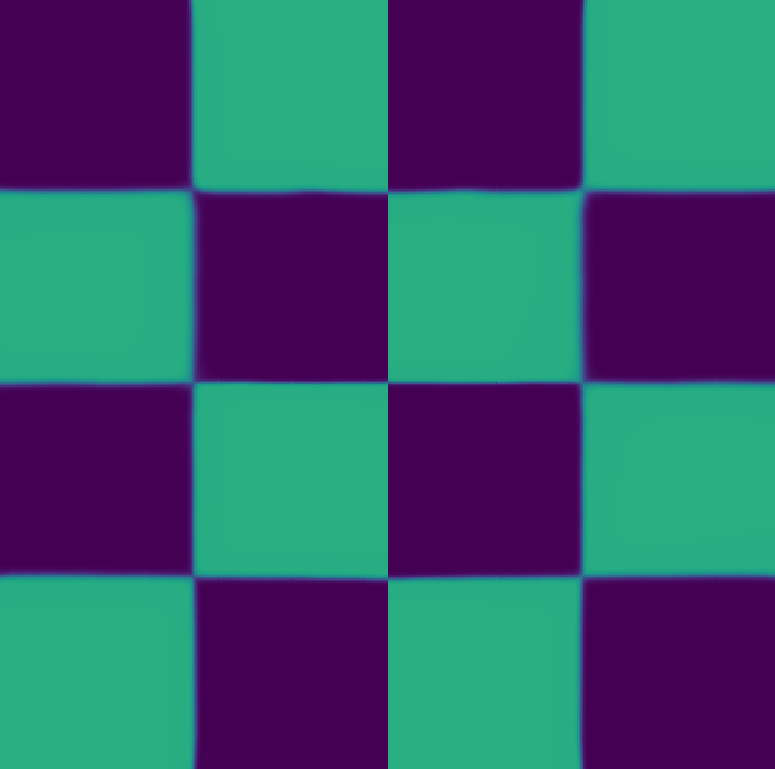}
\end{subfigure}

\begin{subfigure}[b]{\fw}
    \includegraphics[width=\textwidth]{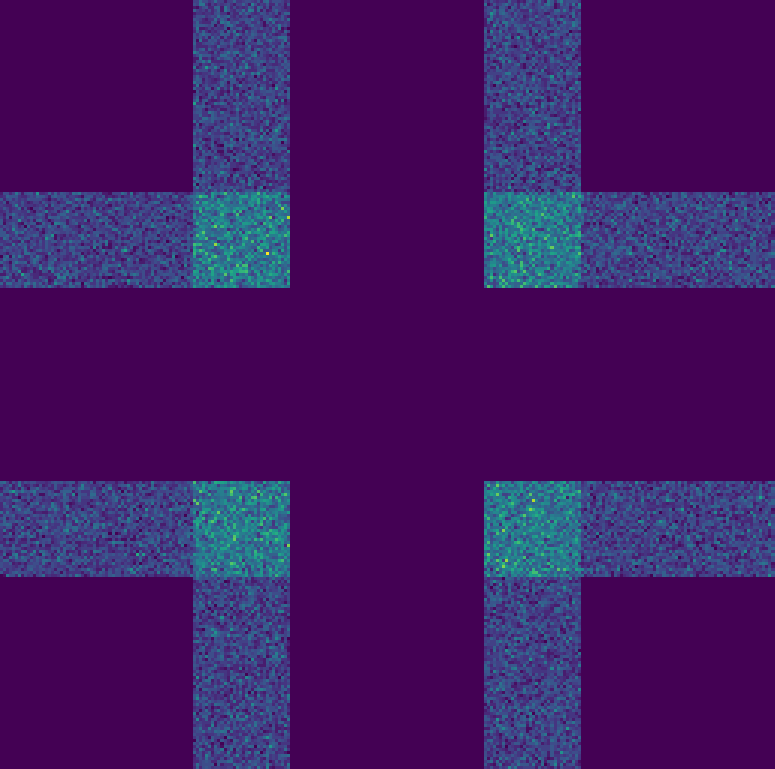}
\end{subfigure}
\begin{subfigure}[b]{\fw}
    \includegraphics[width=\textwidth]{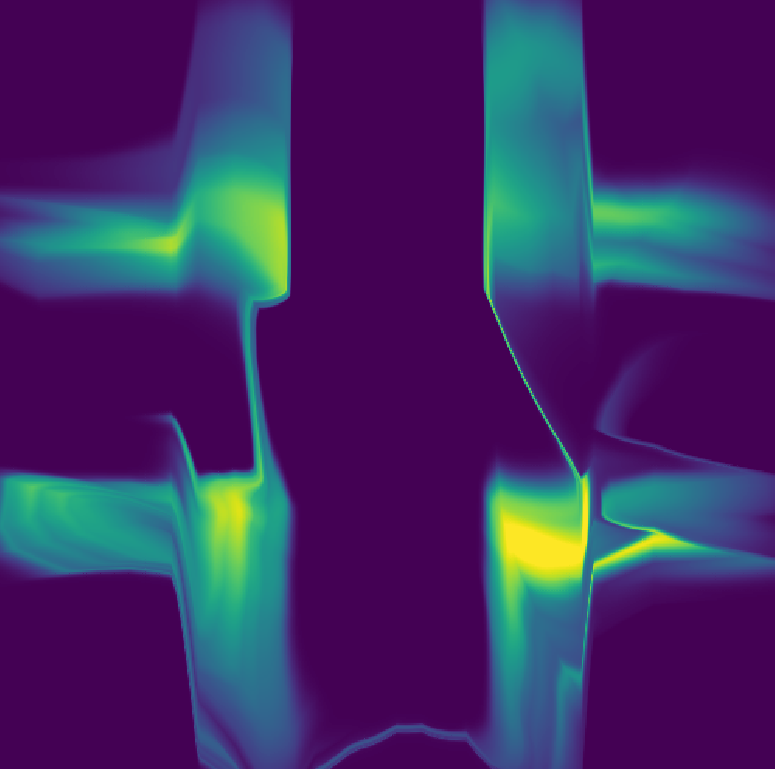}
\end{subfigure}
\begin{subfigure}[b]{\fw}
    \includegraphics[width=\textwidth]{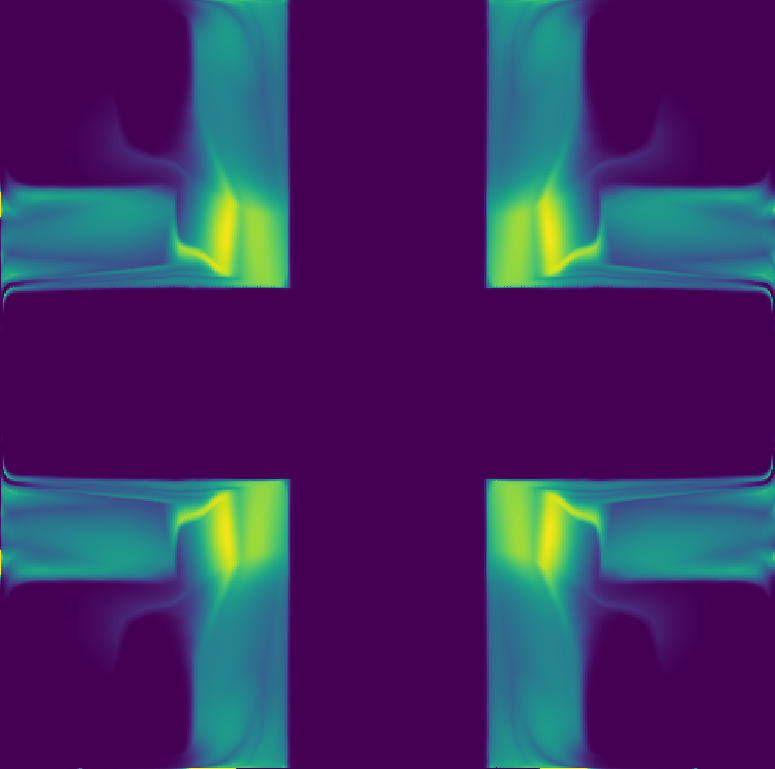}
\end{subfigure}

\begin{subfigure}[b]{\fw}
    \includegraphics[width=\textwidth]{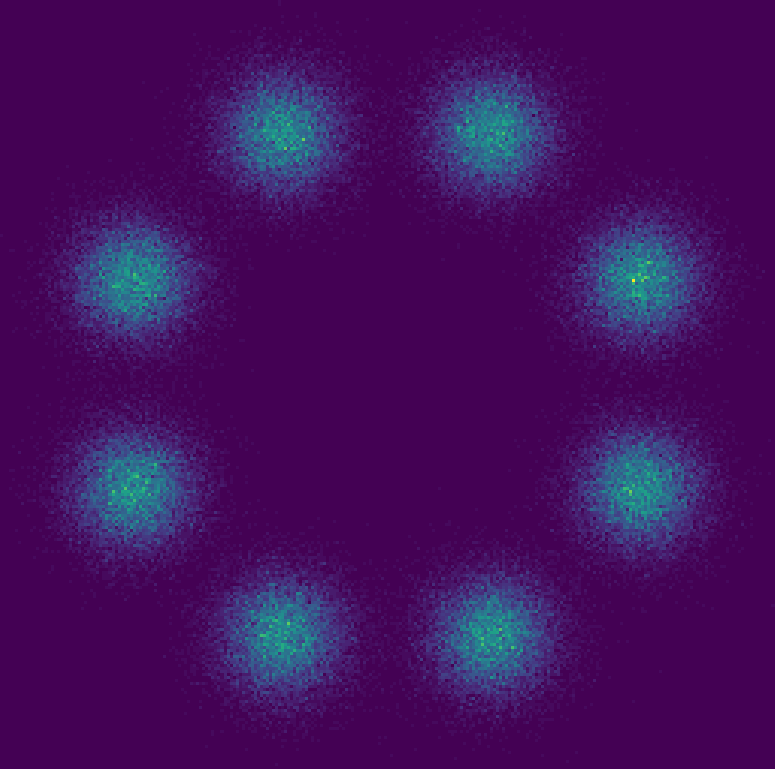}
\end{subfigure}
\begin{subfigure}[b]{\fw}
    \includegraphics[width=\textwidth]{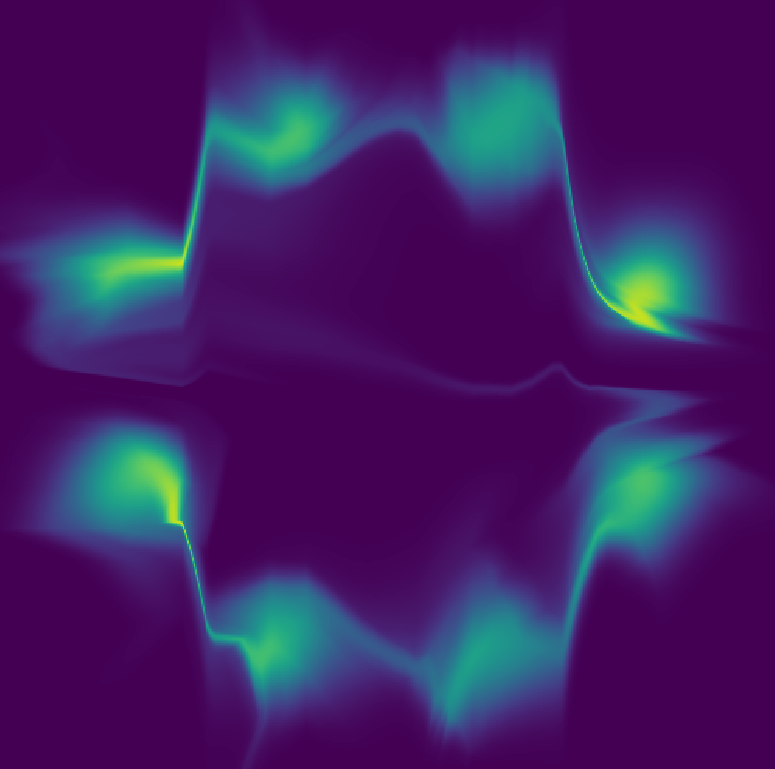}
\end{subfigure}
\begin{subfigure}[b]{\fw}
    \includegraphics[width=\textwidth]{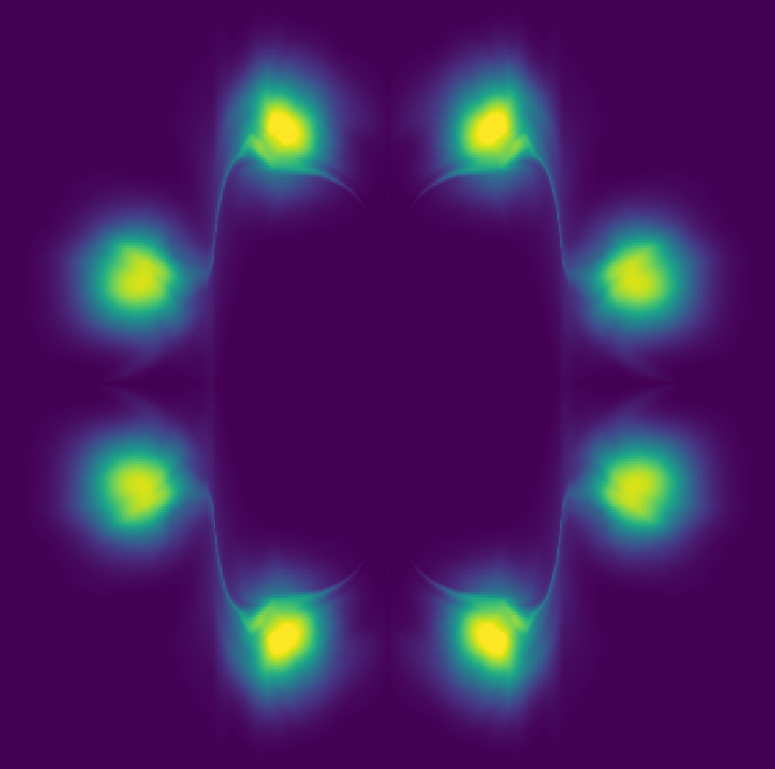}
\end{subfigure}

\begin{subfigure}[b]{\fw}
    \includegraphics[width=\textwidth]{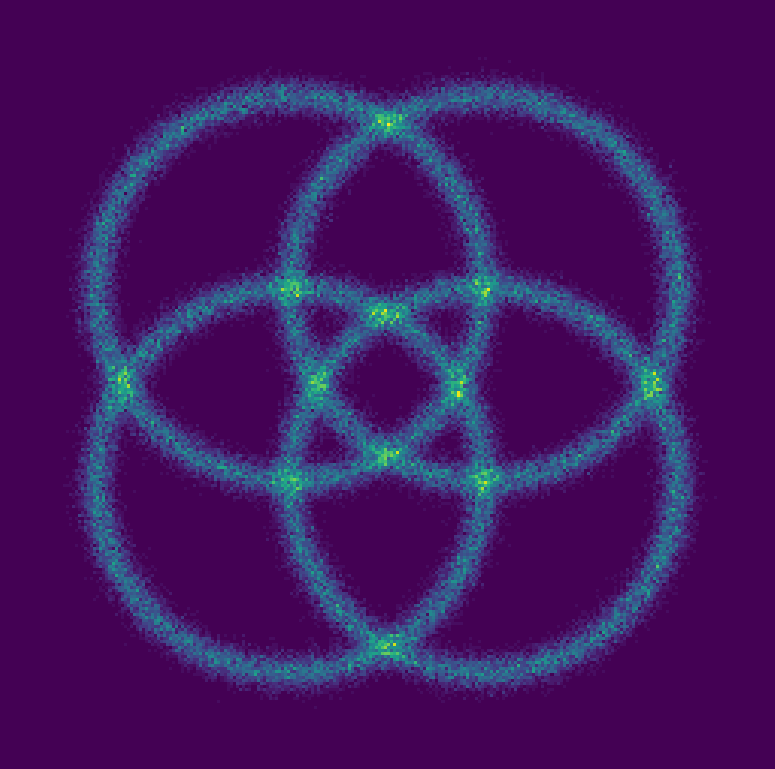}
\end{subfigure}
\begin{subfigure}[b]{\fw}
    \includegraphics[width=\textwidth]{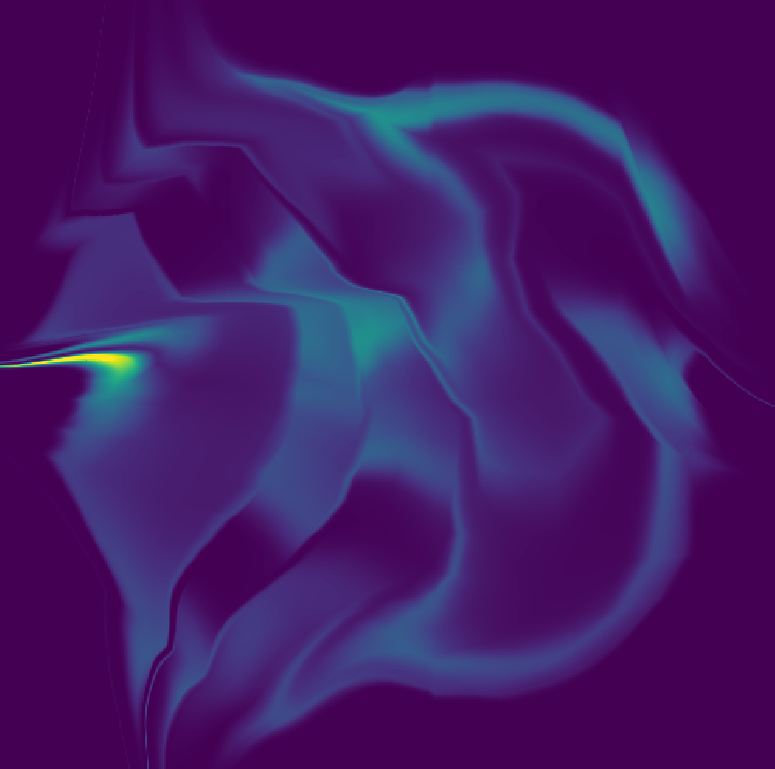}
\end{subfigure}
\begin{subfigure}[b]{\fw}
    \includegraphics[width=\textwidth]{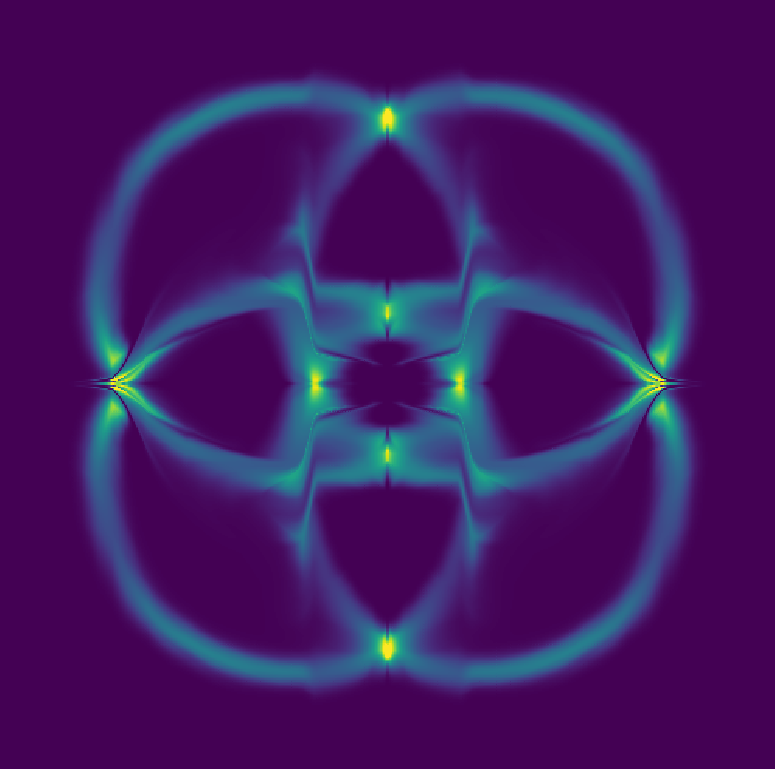}
\end{subfigure}

\begin{tabular}{lcc}
\toprule
\textbf{Dataset} & \textbf{Flow} & \textbf{AbsFlow (ours)} \\
\midrule
Checkerboard & 3.65 & \textbf{3.49} \\
Corners & 3.19 & \textbf{3.03} \\
Gaussians & 3.01 & \textbf{2.86} \\
Circles & 3.44 & \textbf{2.99} \\
\bottomrule
\end{tabular}

\caption{Comparison of flows with and without absolute value surjections modelling anti-symmetric (top row) and symmetric (3 bottom rows) 2-dimensional distributions.}
\label{fig:abs_unif}
\end{wrapfigure}

We investigate the ability of SurVAE flows to model data that is difficult to model with normalizing flows. We show that the absolute value surjection is useful in modelling data where certain symmetries are known to exist. Next, we demonstrate that SurVAE flows allow straightforward modelling of exchangeable data by simply composing \emph{any} flow together with either a sorting surjection or a stochastic permutation layer.
Furthermore, we investigate the use of max pooling -- which is commonly used for downsampling in convolutional neural networks -- as a surjective downsampling layer in SurVAE flows for image data.

\textbf{Synthetic Data.} We first consider modelling data where certain symmetries are known to exist. We make use of 3 symmetric and 1 anti-symmetric synthetic 2D datasets.
The absolute value inference surjection can be seen to fold the input space across the origin and can thus be useful in modelling such data. The baseline uses 4 coupling bijections, while our \emph{AbsFlow} adds an extra \texttt{abs} surjection. For the anti-symmetric data, AbsFlow uses only a single \texttt{abs} surjection with a classifier (\textit{i.e.} for $P(s|z)$) which learns the unfolding.
For further details, see App. \ref{app:synthetic}.
The results are shown in Fig. \ref{fig:abs_unif}.

\newcommand{\fwc}{0.31\linewidth}
\begin{figure}[b]
    \centering

\begin{subfigure}[b]{0.28\textwidth}
    \centering
    \includegraphics[width=\textwidth]{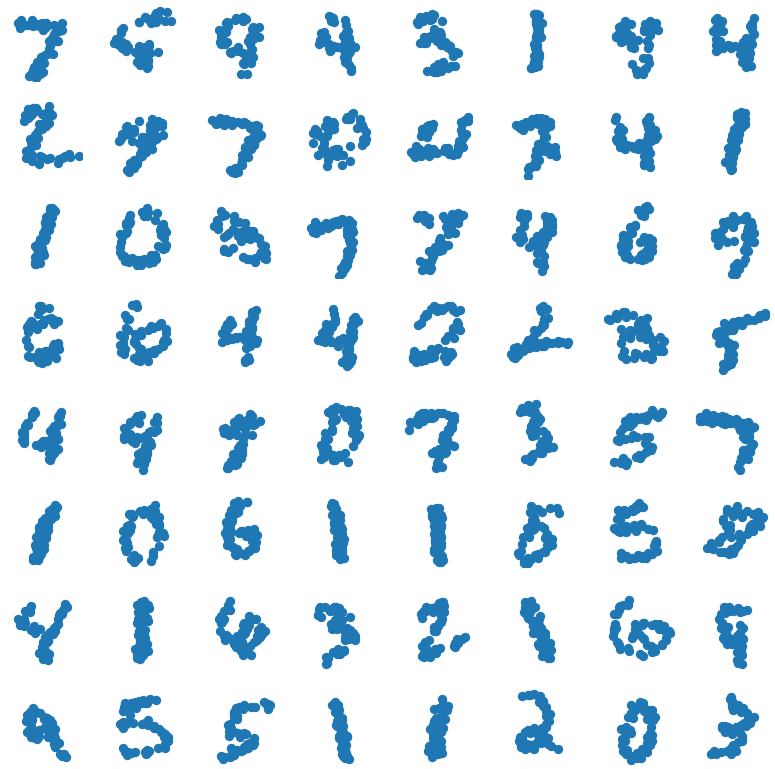}
    \caption{Data}
\end{subfigure}
\quad
\begin{subfigure}[b]{0.28\textwidth}
    \centering
    \includegraphics[width=\textwidth]{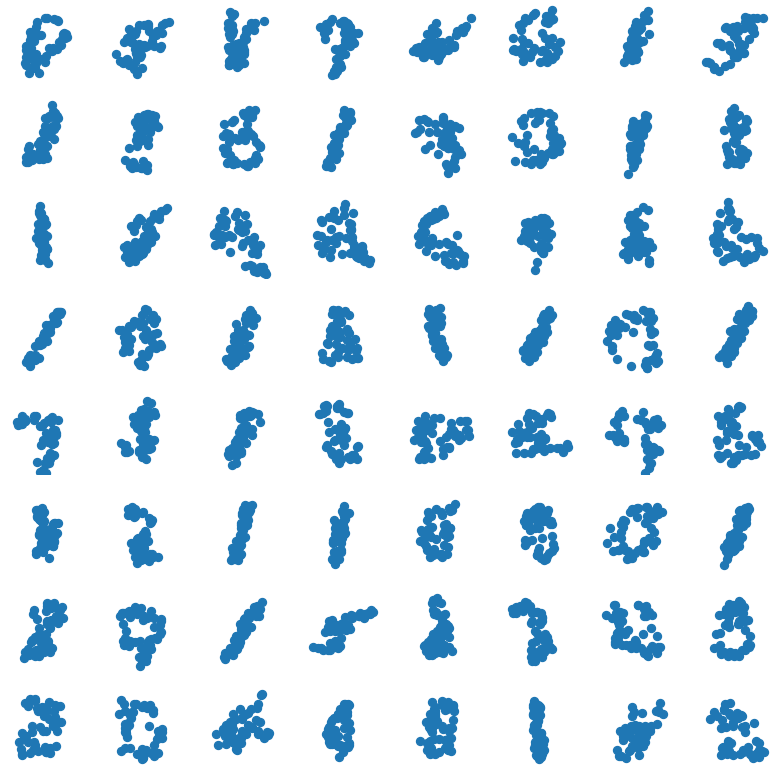}
    \caption{SortFlow}
\end{subfigure}
\quad
\begin{subfigure}[b]{0.28\textwidth}
    \centering
    \includegraphics[width=\textwidth]{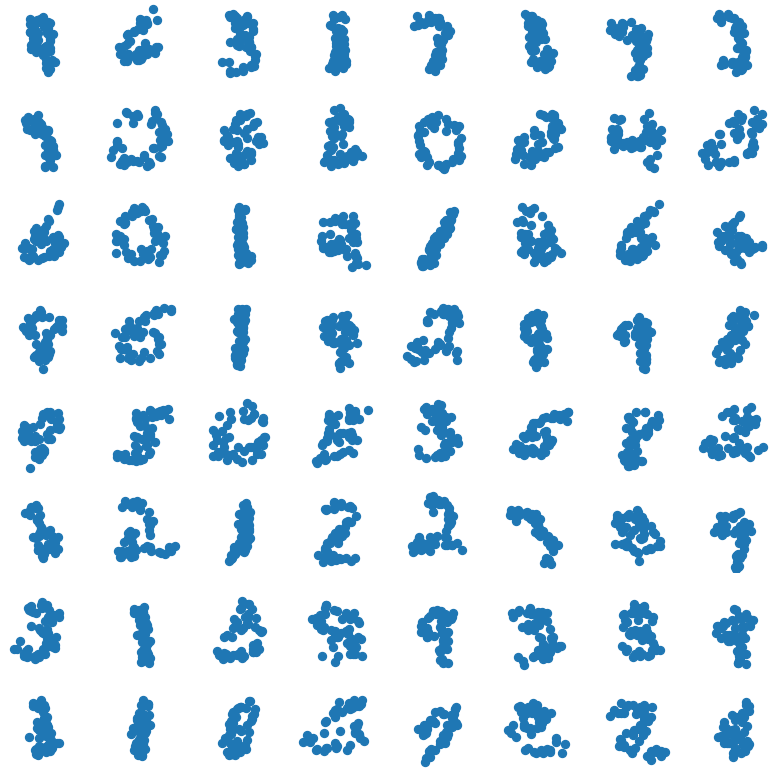}
    \caption{PermuteFlow}
\end{subfigure}

    \caption{Point cloud samples from permutation-invariant SurVAE flows trained on \texttt{SpatialMNIST}.}
    \label{fig:spatial_mnist}
\end{figure}

\textbf{Point Cloud Data.} We now consider modelling exchangeable data.
We use the \texttt{SpatialMNIST} dataset \citep{edwards2017}, where each MNIST digit is represented as a 2D point cloud of 50 points. A point cloud is a set, \textit{i.e.} it is permutation invariant. Using SurVAE flows, we can enforce permutation invariance on \emph{any} flow using either 1) a sorting surjection -- forcing a canonical order on the inputs, or 2) a stochastic permutation -- forcing a random order on the inputs. 

We compare 2 SurVAE flows, \emph{SortFlow} and \emph{PermuteFlow}, both using 64 layers of coupling flows parameterized by Transformer networks \citep{vaswani2017}. Transformers are -- when \emph{not} using positional encoding -- permutation equivariant. PermuteFlow uses stochastic permutation in-between the coupling layers. SortFlow, on the other hand, uses and initial sorting surjection, which introduces an ordering, and fixed permutations after. The Transformers thus make use of learned positional encodings for SortFlow, but not for PermuteFlow. See App. \ref{app:point_cloud} for further details, and Fig. \ref{fig:spatial_mnist} for model samples. Interestingly, PermuteFlow outperforms SortFlow, with -5.30 vs. -5.53 PPLL (per-point log-likelihood), even though it only allows computation of lower bound likelihood estimates.
For comparison, BRUNO \citep{korshunova2018} and FlowScan \citep{bender2019} obtain -5.68 and -5.26 PPLL, but make use of autoregressive components. Neural Statistican \citep{edwards2017} utilizes hierarchical latent variables without autoregressive parts and obtains -5.37 PPL. PermuteFlow thus obtains \emph{state-of-the-art} performance among non-autoregressive models.

\begin{wrapfigure}[17]{r}{0.25\linewidth}
\vspace{-5mm}
\centering
\includegraphics[width=0.9\linewidth]{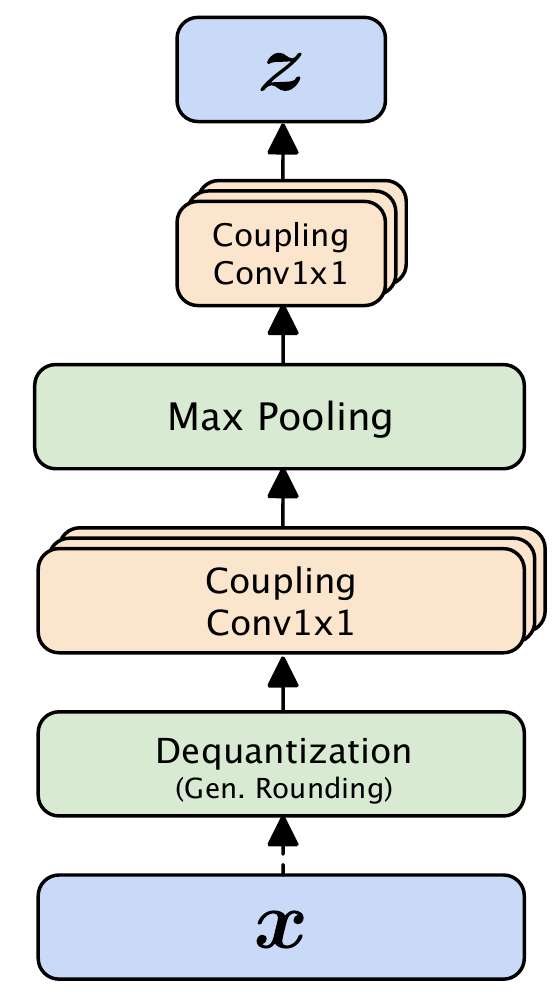}
\caption{Flow architecture with max pooling. Surjections in green.}
\label{fig:arch}
\end{wrapfigure}

\textbf{Image Data.} Max pooling layers are commonly used for downsampling in convolutional neural networks. We investigate their use as surjective downsampling transformations in flow models for image data here.

\begin{figure}[t]
\centering
\begin{minipage}[b]{0.42\textwidth}
\centering
\includegraphics[trim=0 170 0 0,clip,width=0.85\textwidth]{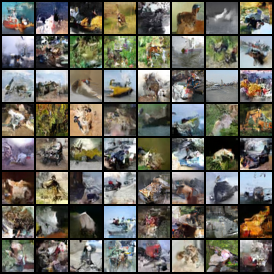}
\includegraphics[trim=0 170 0 0,clip,width=0.85\textwidth]{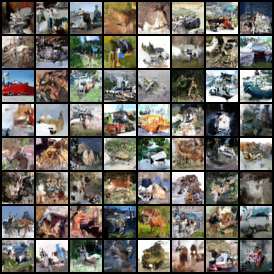}
\caption{Samples from CIFAR-10 models. Top: MaxPoolFlow, Bottom: Baseline.}\label{fig:samples}
\end{minipage}
\hspace{0.5cm}
\begin{minipage}[b]{0.5\textwidth}
\centering
\captionsetup{type=table}
\begin{tabular}{lcc}
\toprule
\textbf{Model} & \textbf{Inception $\uparrow$} & \textbf{FID $\downarrow$} \\
\midrule
DCGAN* & 6.4 & 37.1 \\
WGAN-GP* & 6.5 & 36.4 \\
PixelCNN* & 4.60 & 65.93 \\
PixelIQN* & 5.29 & 49.46 \\
\midrule
Baseline (Ours) & 5.08 & 49.56 \\
MaxPoolFlow (Ours) & \textbf{5.18} & \textbf{49.03} \\
\bottomrule
\end{tabular}
\caption{Inception score and FID for CIFAR-10. \\ *Results taken from \citet{ostrovski2018}.}\label{tab:inception}
\end{minipage}
\end{figure}

We train a flow using 2 scales with 12 steps/scale for \texttt{CIFAR-10} and \texttt{ImageNet 32$\times$32} and 3 scales with 8 steps/scale for \texttt{ImageNet 64$\times$64}. Each step consists of an affine coupling bijection and a 1$\times$1 convolution \citep{kingma2018}. We implement a max pooling surjection for downscaling and compare it to a baseline model with tensor slicing which corresponds to a \emph{multi-scale} architecture \citep{dinh2017}. 
We report results for the log-likelihood in \Cref{tab:results} and the inception and FID scores in \Cref{tab:inception} with bolding indicating best among the baseline and MaxPoolFlow. The results show that compared to slicing surjections, the max pooling surjections yield marginally worse log-likelihoods, but better visual sample quality as measured by the Inception score and FID. We also provide the generated samples from our models in Fig. \ref{fig:samples} and App.~\ref{app:samples}. Due to space constrains, we refer the reader to App. \ref{app:image} for more details on the experiment.

\begin{table}[t]
\centering
\caption{Unconditional image modeling results in bits/dim.}
\label{tab:results}
\begin{tabular}{lccc}
\toprule
\textbf{Model} & \textbf{CIFAR-10} & \textbf{ImageNet32} & \textbf{ImageNet64} \\
\midrule
RealNVP \citep{dinh2017} & 3.49 & 4.28 & - \\
Glow \citep{kingma2018} & 3.35 & 4.09 & 3.81 \\
Flow++ \citep{ho2019} & 3.08 & 3.86 & 3.69 \\
\hline
Baseline (Ours) & \textbf{3.08} & \textbf{4.00} & \textbf{3.70} \\
MaxPoolFlow (Ours) & 3.09 & 4.01 & 3.74 \\
\bottomrule
\end{tabular}
\end{table}

\section{Conclusion}

We introduced SurVAE flows, a modular framework for constructing likelihood-based models using composable bijective, surjective and stochastic transformations. We showed how this encompasses normalizing flows, which rely on bijections, as well as VAEs, which rely on stochastic transformations. We further showed that several recently proposed methods such as dequantization and augmented normalizing flows may be obtained as SurVAE flows using surjective transformations. One interesting direction for further research is development of novel non-bijective transformations that might be beneficial as composable layers in SurVAE flows.

\section*{Acknowledgements}
We thank Laurent Dinh and Giorgio Giannone for helpful feedback.

\section*{Funding Disclosure}
This research was supported by the NVIDIA Corporation with the donation of TITAN X GPUs.

\section*{Broader Impact}

This work constitutes foundational research on generative models/unsupervised learning by providing a unified view on several lines of work and further by introducing new modules that expand the generative modelling toolkit. This work further suggests how to build software libraries to that allows more rapid implementation of a wider range of deep unsupervised models.
Unsupervised learning has the potential to greatly reduce the need for labeled data and thus improve models in applications such as medical imaging where a lack of data can be a limitation. 
However, it may also potentially be used to improve deep fakes with potentially malicious applications.

\bibliographystyle{apalike}
\bibliography{references}

\newpage
\appendix

\section{A Connection Between VAEs and Flows}
\label{app:vae_limit}

Variational Autoencoders (VAEs) can be seen as a composable stochastic transformations. From this viewpoint, the log-likelihood resulting from a single transformation can be written as
\begin{equation}
    \label{eq:logpx_vae}
    \log p(\vx) = \bE_{q(\vz|\vx)} \left[ \log p(\vz) \right] + \underbrace{\bE_{q(\vz|\vx)} \left[ \log \frac{p(\vx|\vz)}{q(\vz|\vx)} \right]}_{\mathrm{Lik. \,contrib.}\, \cV(\vx, \vz)} + \underbrace{\bE_{q(\vz|\vx)} \left[ \log \frac{q(\vz|\vx)}{p(\vz|\vx)} \right]}_{\mathrm{Bound \,looseness}\, \cE(\vx, \vz)},
\end{equation}
which consists of \textit{1)} the log-likelihood of $\vz \sim q(\vz|\vx)$ under the remaining layers $p(\vz)$, \textit{2)} the likelihood contribution term $\cV(\vx, \vz)$ and \textit{3)} the looseness of the bound $\cE(\vx, \vz)$.

Normalizing flows, on the other hand, make use of deterministic transformations. Specifically, using a diffeomorphism $f: \mathcal{Z} \to \mathcal{X}$, the log-likelihood can be computed as
\begin{equation}
    \label{eq:logpx_flow}
    \log p(\vx) = \log p(\vz) + \underbrace{\log \left|\det \frac{\partial \vz}{\partial \vx}\right|}_{\mathrm{Lik. \,contrib.}\, \cV(\vx, \vz)}, \quad \vz = f^{-1}(\vx),
\end{equation}
which consists \textit{1)} the log-likelihood of $\vz = f^{-1}(\vx)$ under $p(\vz)$ (possibly another flow) and \textit{2)} the likelihood contribution term, which here corresponds to the log Jacobian determinant. Notice that for normalizing flows, the likelihood is exact and hence the \textit{bound looseness} term $\cE(\vx,\vz)=0$.

In the remainder of this section we show that the change-of-variables formula (Eq. \ref{eq:logpx_flow}) can be obtained from the ELBO (Eq. \ref{eq:logpx_vae}).  

\textbf{Proof.} We can use a composition of a function $g$ with a Dirac $\delta$-function:
\begin{equation}
    \int \delta(g(\vz)) f(g(\vz)) \left|\det \frac{\partial g(\vz)}{\partial \vz}\right| d\vz = \int \delta(\vu) f(\vu) d\vu
\end{equation}
to conclude that
\begin{equation}
    \delta(g(\vz)) = \left|\det \frac{\partial g(\vz)}{\partial \vz}\right|_{\vz=\vz_0}^{-1} \delta(\vz-\vz_0)
\end{equation}
with $\vz_0$ being the root of $g(\vz)$. This results assumes that $g$ is smooth (derivative exists), $f$  has compact support, the root is unique and the Jacobian is non-singular.

Let $f: \mathcal{Z} \to \mathcal{X}$ be a diffeomorphism and define a pair of deterministic conditionals
\begin{align}
    p(\vx|\vz) &= \delta(\vx-f(\vz)) \\
    p(\vz|\vx) &= \delta(\vz-f^{-1}(\vx)).
\end{align}
Applying the above result to $p(\vx|\vz)$, we set $g(\vz)=\vx-f(\vz)$ and find $\vz_0=f^{-1}(\vx)$ and
\begin{equation}
    p(\vx|\vz) = \delta(\vz-f^{-1}(\vx)) |\det \vJ| = p(\vz|\vx) |\det \vJ|\ ,
\end{equation}
where
$$
\vJ^{-1} = \left. \frac{\partial f(\vz)}{\partial \vz} \right|_{\vz=f^{-1}(\vx)} \ . 
$$
Let further $q(\vz|\vx) = p(\vz|\vx) = \delta(\vz-f^{-1}(\vx))$. The resulting ELBO gives rise to the change-of-variables formula,
\begin{align}
    \log p(\vx) &= \bE_{q(\vz|\vx)} \left[  \log p(\vz) + \log \frac{p(\vx|\vz)}{q(\vz|\vx)} + \log \frac{q(\vz|\vx)}{p(\vz|\vx)} \right] \\
    &= \log p(\vz) + \log |\det \vJ|, \quad \mathrm{for}\,\, \vz=f^{-1}(\vx),
\end{align}
where the likelihood contribution $\cV(\vx,\vz) = \log \frac{p(\vx|\vz)}{q(\vz|\vx)} = \log |\det \vJ|$, while the bound looseness term $\cE(\vx,\vz) = \log \frac{q(\vz|\vx)}{p(\vz|\vx)} = 0$, trivially.

\section{The Bound Looseness for Inference Surjections}
\label{app:bound_looseness}

For inference surjections $f: \cX \rightarrow \cZ$, the bound looseness term $\cE(\vx,\vz) = 0$, given that the \emph{stochastic right inverse condition} is satified. 
The stochastic right inverse condition requires that $p(\vx|\vz)$ defines a distribution over the possible right inverses of the surjection $f$. 

A right inverse function $g: \cZ \rightarrow \cX$ to a function $f: \cX \rightarrow \cZ$ satisfies $f \circ g = \mathrm{id}_{\cZ}$, 
but not necessarily $g \circ f = \mathrm{id}_{\cX}$. Here $\mathrm{id}_{\cS}$ denotes an identity map defined on the space $\cS$.

We satisfy the stochastic right inverse condition by requiring that $p(\vx|\vz)$ only has support over the \emph{fiber} of $\vz$, i.e. the set of elements $\cB(\vz)$ in the domain $\cX$ that are mapped to $\vz$, $\cB(\vz) := \{\vx | \vz = f(\vx)\}$. 
A simple check for stochastic right invertibility is thus: For any $\vz$, computing $\vz = f(\vx)$, for $\vx \sim p(\vx|\vz)$ should return the original $\vz$.

Given that the distribution $p(\vz)$ has full support over $\cZ$ and the stochastic right inverse condition is satisfied, we have that, for any observed $\vx$, only one $\vz$ could have given rise to the observation $\vx$. Consequently, the posterior distribution $p(\vz|\vx) = \delta(\vz - f(\vx))$ is deterministic. By defining $q(\vz|\vx) = p(\vz|\vx)$, the bound looseness is thus $\cE(\vx,\vz) = 0$.

\section{List of SurVAE Layers}
\label{app:layers}

See Table \ref{tab:generative_surjections} and Table \ref{tab:inference_surjections} for lists of generative and inference surjection layers, respectively. 

\begin{table}[h]
\centering
\caption{Summary of some generative surjection layers.}
\label{tab:generative_surjections}
\footnotesize
\begin{tabular}{l|c|c|c}
\toprule
\textbf{Surjection} & \textbf{Forward} & \textbf{Inverse} & $\cV(\vx, \vz)$ \\ \toprule
Rounding & $x = \lfloor z \rfloor$ & $z \sim q(z|x)$ where $z \in [x,x+1)$ & $-\log q(z|x)$ \\ \midrule
Slicing & $\vx = \vz_1$ & $\vz_1 = \vx, \vz_2 \sim q(\vz_2|\vx)$ & $-\log q(\vz_2|\vx)$ \\ \midrule
\multirow{2}{*}{Abs} & $s = \sign z$ & $s \sim \Bern(\pi(x))$ & \multirow{2}{*}{$- \log q(s|x)$} \\
 & $x = |z|$ & $z = s\cdot x, ~~s \in \{1, -1\}$ &  \\ \midrule
\multirow{2}{*}{Max} & $k = \argmax \vz$ & $k \sim \Cat(\vpi(x))$ & \multirow{2}{*}{$- \log q(k|x) - \log q(\vz_{-k} | x, k)$} \\
 & $x = \max \vz$ & $z_k = x, \vz_{-k}  \sim q(\vz_{-k} | x, k)$ &  \\ \midrule
\multirow{2}{*}{Sort} & $\cI = \argsort \vz$ & $\cI \sim \Cat(\vpi(\vx))$ & \multirow{2}{*}{$-\log q(\cI|\vx)$} \\
 & $\vx = \sort \vz$ & $\vz = \vx_{\cI}$ &  \\ 
 \midrule
ReLU & $x = \max(z, 0)$ & $\mathrm{if}\, x=0: z \sim q(z), \mathrm{else}: z=x$ & $\bI(x = 0) [-\log q(z)]$ \\ 
\bottomrule
\end{tabular}
\end{table}
\normalsize

\begin{table}[h]
\caption{Summary of some inference surjection layers.}
\label{tab:inference_surjections}
\footnotesize
\centering
\begin{tabular}{l|c|c|c}
\toprule
\textbf{Surjection} & \textbf{Forward} & \textbf{Inverse} & $\cV(\vx, \vz)$ \\ \midrule
Rounding & $x \sim p(x|z)$ where $x \in [z,z+1)$ & $z = \lfloor x \rfloor$ & $\log p(z|x)$ \\ \midrule
Slicing & $\vx_1 = \vz, \vx_2 \sim p(\vx_2|\vz)$ & $\vz = \vx_1$ & $\log p(\vx_2|\vz)$ \\ \midrule
\multirow{2}{*}{Abs} & $s \sim \Bern(\pi(z))$ & $s = \sign x$ & \multirow{2}{*}{$\log p(s|z)$} \\
 &  $x = s \cdot z, ~~ s \in \{-1, 1\}$ & $z = |x|$ &  \\ \midrule
\multirow{2}{*}{Max} & $k \sim \Cat(\vpi(z))$ & $k=\argmax \vx$ & \multirow{2}{*}{$\log p(k|z) + \log p(\vx_{-k} | z, k)$} \\
 &  $x_k = z, \vx_{-k}  \sim p(\vx_{-k} | z, k)$ & $z = \max \vx$ &  \\ \midrule
\multirow{2}{*}{Sort} & $\cI \sim \Cat(\vpi(\vz))$ & $\cI = \argsort \vx$ & \multirow{2}{*}{$\log p(\cI|\vz)$} \\
 & $\vx = \vz_{\cI}$ & $\vz = \sort \vx$ &  \\ 
 \midrule
ReLU & $\mathrm{if}\, z=0: x \sim p(x), \mathrm{else}: x=z$ & $z = \max(x, 0)$ & $\bI(z = 0) \log p(x)$ \\ 
\bottomrule
\end{tabular}
\end{table}
\normalsize

\newpage
\section{The Absolute Value Surjection}
\label{app:abs}

We here develop the absolute value surjections, both in the generative direction $x = |z|$ and in the inference direction $z = |x|$. We will make use of Dirac delta functions to develop the likelihood contributions, but we could equivalently develop them using Gaussian distributions where $\sigma \rightarrow 0$.

\subsection{Generative Direction}

\textbf{Forward and Inverse.} We define the forward and inverse transformations as
\begin{align}
    p(x|z) &= \sum_{s\in\{-1,1\}}p(x|z,s)p(s|z) = \sum_{s\in\{-1,1\}}\delta(x-sz)\delta_{s,\sign(z)}, \\
    q(z|x) &= \sum_{s\in\{-1,1\}}q(z|x,s)q(s|x) = \sum_{s\in\{-1,1\}}\delta(z-sx)q(s|x),
\end{align}
where the forward transformation $p(x|z)$ is fully deterministic and corresponds to $x=|z|$. The inference direction involves two steps, 1) sample the sign $s$ of $z$ conditioned of $x$, and 2) deterministically map $x$ to $z=sx$. Note that $q(s|x)$ may either be trained as a classifier or fixed to e.g. $q(s|x) = 1/2$. The last choice especially makes sense when $p(z)$ is symmetric.

\textbf{Likelihood Contribution.} We may develop the likelihood contribution by computing 
\begin{align}
    \cV &= \bE_{q(z|x,s)q(s|x)} \left[ \log \frac{p(x|z,s)p(s|z)}{q(z|x,s)q(s|x)} \right] \\
    &= \bE_{\delta(z-sx)q(s|x)} \left[ \log \frac{\delta(x-sz)\delta_{s,\sign(z)}}{\delta(z-sx)q(s|x)} \right] \\
    &\approx - \log q(s|x), \quad\mathrm{where}\,\, z = sx,\,\, s \sim q(s|x).
\end{align}
Here, $\delta(x-sz)$ and $\delta(z-sx)$ cancel since $\delta(x-sz) = \delta(z-x/s) |1/s| = \delta(z-sx)$.

\subsection{Inference Direction}

\textbf{Forward and Inverse.} We define the forward and inverse transformations as
\begin{align}
    p(x|z) &= \sum_{s\in\{-1,1\}}p(x|z,s)p(s|z) = \sum_{s\in\{-1,1\}}\delta(x-sz)p(s|z), \\
    q(z|x) &= \sum_{s\in\{-1,1\}}q(z|x,s)q(s|x) = \sum_{s\in\{-1,1\}}\delta(z-sx)\delta_{s,\sign(x)},
\end{align}
where the inverse transformation $q(z|x)$ is fully deterministic and corresponds to $z=|x|$. The generative direction involves two steps, 1) sample the sign of $x$ conditioned of $z$, and 2) deterministically map $z$ to $x=sz$. Note that $p(s|z)$ may either be trained as a classifier or fixed to e.g. $p(s|z) = 1/2$. The last choice gives rise to an absolute value surjection which may be used to enforce exact symmetry across the origin.

\textbf{Likelihood Contribution.} We may develop the likelihood contribution by computing 
\begin{align}
    \cV &= \bE_{q(z|x,s)q(s|x)} \left[ \log \frac{p(x|z,s)p(s|z)}{q(z|x,s)q(s|x)} \right] \\
    &= \bE_{\delta(z-sx)\delta_{s,\sign(x)}} \left[ \log \frac{\delta(x-sz)p(s|z)}{\delta(z-sx)\delta_{s,\sign(x)}} \right] \\
    &= \log p(s|z), \quad\mathrm{where}\,\, z = sx = |x|,\,\, s = \sign(x).
\end{align}
Here, $\delta(x-sz)$ and $\delta(z-sx)$ cancel since $\delta(x-sz) = \delta(z-x/s) |1/s| = \delta(z-sx)$.

\newpage
\section{The Maximum Value Surjection}
\label{app:max}

We here develop the maximum value surjections, both in the generative direction $x = \max \vz$ and in the inference direction $z = \max \vx$. We will make use of Dirac delta functions to develop the likelihood contributions, but we could equivalently develop them using Gaussian distributions where $\sigma \rightarrow 0$.

\subsection{Generative Direction}

\textbf{Forward and Inverse.} We define the forward and inverse transformations as
\begin{align}
    p(x|\vz) &= \sum_{k=1}^K p(x|\vz,k)p(k|\vz) = \sum_{k=1}^K \delta(x-z_k)\delta_{k,\argmax(\vz)}, \\
    q(\vz|x) &= \sum_{k=1}^K q(\vz|x,k)q(k|x) = \sum_{k=1}^K \delta(z_k-x)q(\vz_{-k}|x,k)q(k|x),
\end{align}
where $k$ refers to the indices of $\vz$, $K$ is the number of elements in $\vz$ and $\vz_{-k}$ is $\vz$ excluding element $k$. 
The forward transformation $p(x|\vz)$ is fully deterministic and corresponds to $x=\max \vz$. The inference direction involves three steps, 1) sample the index $k$ for the argmax of $\vz$ conditioned of $x$, 2) deterministically map $x$ to $z_k=x$, and 3) infer the remaining elements $\vz_{-k}$ of $\vz$. 
Note that $q(k|x)$ may either be trained as a classifier or fixed to e.g. $q(k|x) = 1/K$.

For $q$ to define a right-inverse of $p$, we require that $q(\vz_{-k}|x,k)$ only has support in $(-\infty, x)^{K-1}$ such that $z_k$ will be the maximum value.

\textbf{Likelihood Contribution.} We may develop the likelihood contribution by computing 
\begin{align}
    \cV &= \bE_{q(\vz|x,k)q(k|x)} \left[ \log \frac{p(x|\vz,k)p(k|\vz)}{q(\vz|x,k)q(k|x)} \right] \\
    &= \bE_{\delta(z_k-x)q(\vz_{-k}|x,k)q(k|x)} \left[ \log \frac{\delta(x-z_k)\delta_{k,\argmax(\vz)}}{\delta(z_k-x)q(\vz_{-k}|x,k)q(k|x)} \right] \\
    &\approx - \log q(k|x) - \log q(\vz_{-k}|x,k), \quad\mathrm{where}\,\, z_k = x,\,\, \vz_{-k} \sim q(\vz_{-k}|x,k),\,\, k \sim q(k|x).
\end{align}

\subsection{Inference Direction}

\textbf{Forward and Inverse.} We define the forward and inverse transformations as
\begin{align}
    p(\vx|z) &= \sum_{k=1}^K p(\vx|z,k)p(k|z) = \sum_{k=1}^K \delta(x_k-z)p(\vx_{-k}|z,k)p(k|z), \\
    q(z|\vx) &= \sum_{k=1}^K q(z|\vx,k)q(k|x) = \sum_{k=1}^K \delta(z-x_k)\delta_{k,\argmax(\vx)},
\end{align}
where $k$ refers to the indices of $\vx$, $K$ is the number of elements in $\vx$ and $\vx_{-k}$ is $\vx$ excluding element $k$. 
The inverse transformation $q(z|\vx)$ is fully deterministic and corresponds to $z=\max \vx$. The inference direction involves three steps, 1) sample the index $k$ for the argmax of $\vx$ conditioned of $z$, 2) deterministically map $z$ to $x_k=z$, and 3) infer the remaining elements $\vx_{-k}$ of $\vx$. 
Note that $p(k|z)$ may either be trained as a classifier or fixed to e.g. $p(k|z) = 1/K$.

For $p$ to define a right-inverse of $q$, we require that $p(\vx_{-k}|z,k)$ only has support in $(-\infty, z)^{K-1}$ such that $x_k$ will be the maximum value.

\textbf{Likelihood Contribution.} We may develop the likelihood contribution by computing 
\begin{align}
    \cV &= \bE_{q(z|\vx,k)q(k|\vx)} \left[ \log \frac{p(\vx|z,k)p(k|z)}{q(z|\vx,k)q(k|\vx)} \right] \\
    &= \bE_{\delta(z-x_k)\delta_{k,\argmax(\vx)}} \left[ \log \frac{\delta(x_k-z)p(\vx_{-k}|z,k)p(k|z)}{\delta(z-x_k)\delta_{k,\argmax(\vx)}} \right] \\
    &= \log p(k|z) + \log p(\vx_{-k}|z,k), \quad\mathrm{where}\,\, z = x_k = \max \vx,\,\, k = \argmax \vx.
\end{align}

\newpage
\section{The Sort Surjection}
\label{app:sort}

We here develop the sorting surjections, both in the generative direction $x = \sort \vz$ and in the inference direction $z = \sort \vx$. We will make use of Dirac delta functions to develop the likelihood contributions, but we could equivalently develop them using Gaussian distributions where $\sigma \rightarrow 0$.

\subsection{Generative Direction}

\textbf{Forward and Inverse.} We define the forward and inverse transformations as
\begin{align}
    p(\vx|\vz) &= \sum_{\cI} p(\vx|\vz,\cI)p(\cI|\vz) = \sum_{\cI} \delta(\vx-\vz_{\cI})\delta_{\cI,\argsort(\vz)}, \\
    q(\vz|\vx) &= \sum_{\cI} q(\vz|\vx,\cI)q(\cI|\vx) = \sum_{\cI} \delta(\vz-\vx_{\cI^{-1}})q(\cI|\vx),
\end{align}
where $\cI$ refers to a set of permutation indices, $\cI^{-1}$ refers to the inverse permutation indices and $\vz_{\cI}$ refers to the elements of $\vz$ permuted according to the indices $\cI$. Note that there are $D!$ possible permutations.

The forward transformation $p(\vx|\vz)$ is fully deterministic and corresponds to $\vx=\sort \vz$. The inference direction involves two steps, 1) sample permutation indices $\cI$ conditioned of $\vx$, and 2) deterministically permute $\vx$ according to the inverse permutation $\cI^{-1}$ to obtain $\vz=\vx_{\cI^{-1}}$. 
Note that $q(\cI|\vx)$ may either be trained as a classifier or fixed to e.g. $q(\cI|\vx) = 1/D!$.

\textbf{Likelihood Contribution.} We may develop the likelihood contribution by computing 
\begin{align}
    \cV &= \bE_{q(\vz|\vx,\cI)q(\cI|\vx)} \left[ \log \frac{p(\vx|\vz,\cI)p(\cI|\vz)}{q(\vz|\vx,\cI)q(\cI|\vx)} \right] \\
    &= \bE_{\delta(\vz-\vx_{\cI^{-1}})q(\cI|\vx)} \left[ \log \frac{\delta(\vx-\vz_{\cI})\delta_{\cI,\argsort(\vz)}}{\delta(\vz-\vx_{\cI^{-1}})q(\cI|\vx)} \right] \\
    &\approx - \log q(\cI|\vx), \quad\mathrm{where}\,\, \cI \sim q(\cI|\vx).
\end{align}

\subsection{Inference Direction}

\textbf{Forward and Inverse.} We define the forward and inverse transformations as
\begin{align}
    p(\vx|\vz) &= \sum_{\cI} p(\vx|\vz,\cI)p(\cI|\vz) = \sum_{\cI} \delta(\vx-\vz_{\cI^{-1}})p(\cI|\vz), \\
    q(\vz|\vx) &= \sum_{\cI} q(\vz|\vx,\cI)q(\cI|\vx) = \sum_{\cI} \delta(\vz-\vx_{\cI})\delta_{\cI,\argsort(\vx)},
\end{align}
where $\cI$ refers to a set of permutation indices, $\cI^{-1}$ refers to the inverse permutation indices and $\vx_{\cI}$ refers to the elements of $\vx$ permuted according to the indices $\cI$. Note that there are $D!$ possible permutations.

The inverse transformation $q(\vz|\vx)$ is fully deterministic and corresponds to $\vz=\sort \vx$. The generative direction involves two steps, 1) sample permutation indices $\cI$ conditioned of $\vz$, and 2) deterministically permute $\vz$ according to the inverse permutation $\cI^{-1}$ to obtain $\vx=\vz_{\cI^{-1}}$. 
Note that $p(\cI|\vz)$ may either be trained as a classifier or fixed to e.g. $p(\cI|\vz) = 1/D!$.

\textbf{Likelihood Contribution.} We may develop the likelihood contribution by computing 
\begin{align}
    \cV &= \bE_{q(\vz|\vx,\cI)q(\cI|\vx)} \left[ \log \frac{p(\vx|\vz,\cI)p(\cI|\vz)}{q(\vz|\vx,\cI)q(\cI|\vx)} \right] \\
    &= \bE_{\delta(\vz-\vx_{\cI})\delta_{\cI,\argsort(\vx)}} \left[ \log \frac{\delta(\vx-\vz_{\cI^{-1}})p(\cI|\vz)}{\delta(\vz-\vx_{\cI})\delta_{\cI,\argsort(\vx)}} \right] \\
    &= \log p(\cI|\vz), \quad\mathrm{where}\,\, \vz = \vx_{\cI} = \sort \vx,\,\, \cI = \argsort \vx.
\end{align}

\newpage
\section{The Stochastic Permutation}
\label{app:permute}

We here develop the stochastic permutation layer which randomly permutes its input. The inverse pass mirrors the forward pass. Note that stochastic permutation is \emph{not} a surjection, but rather a stochastic transform. We will make use of Dirac delta functions to develop the likelihood contributions, but we could equivalently develop them using Gaussian distributions where $\sigma \rightarrow 0$.

\textbf{Forward and Inverse.} We define the forward and inverse transformations as
\begin{align}
    p(\vx|\vz) &= \sum_{\cI} p(\vx|\vz,\cI)p(\cI) = \sum_{\cI} \delta(\vx-\vz_{\cI})\Unif(\cI), \\
    q(\vz|\vx) &= \sum_{\cI} q(\vz|\vx,\cI)q(\cI) = \sum_{\cI} \delta(\vz-\vx_{\cI^{-1}})\Unif(\cI),
\end{align}
where $\cI$ refers to a set of permutation indices, $\cI^{-1}$ refers to the inverse permutation indices and $\vz_{\cI}$ refers to the elements of $\vz$ permuted according to the indices $\cI$. Note that there are $D!$ possible permutations.

The transformation is stochastic and involves the same two steps in both directions: 1) Sample permutation indices $\cI$ uniformly at random, and 2) deterministically permute the input according to the samples indices $\cI$.

\textbf{Likelihood Contribution.} We may develop the likelihood contribution by computing 
\begin{align}
    \cV &= \bE_{q(\vz|\vx,\cI)q(\cI)} \left[ \log \frac{p(\vx|\vz,\cI)p(\cI)}{q(\vz|\vx,\cI)q(\cI)} \right] \\
    &= \bE_{\delta(\vz-\vx_{\cI^{-1}})\Unif(\cI)} \left[ \log \frac{\delta(\vx-\vz_{\cI})\Unif(\cI)}{\delta(\vz-\vx_{\cI^{-1}})\Unif(\cI)} \right] \\
    &= 0.
\end{align}
This layer thus takes the simple form: Both during the forward and inverse passes, shuffle the input uniformly at random. The resulting likelihood contribution is zero.

\newpage
\section{The Software Perspective}
\label{app:software}

Normalizing flows provide a powerful modular framework where flexible densities may be specified using a composition of bijective transformations. Each bijection may be implemented as a module contained 3 important components: 1) A forward transformation $\vx = f(\vz)$, 2) an inverse transformation $\vz = f^{-1}(\vx)$, and 3) a Jacobian determinant $\log |\det \vJ|$. Several software libraries for normalizing flows have been built using this modular design principle \citep{dillon2017, bingham2018}.

SurVAE flows suggest that such software frameworks may be directly extended since the modules follow the exact same design principles -- each module has 3 important components:
\begin{enumerate}
    \item A forward transformation $\cZ \rightarrow \cX$.
    \item An inverse transformation $\cX \rightarrow \cZ$.
    \item A likelihood contribution $\cV(\vx,\vz)$.
\end{enumerate}
SurVAE flows allow compositions of not only bijective transformations, but also surjective and stochastic transformations. This allows us to obtain methods such as dequantization \citep{uria2014, theis2016, ho2019}, variational data augmentation \citep{huang2020, chen2020}, multi-scale architectures \citep{dinh2017} as composable surjective transformations and VAEs \citep{kingma2013, rezende2014} as composable stochastic transformations.

In our code\footnote{The code is available at \url{https://github.com/didriknielsen/survae_flows}}, we provide a library of SurVAE flows that may serve as a prototype for a more extensive library. In the next subsections, we show some selected code snippets from our library. The code is based on PyTorch \citep{paszke2019}, but can easily be ported to other frameworks. Note that in the implementation, the \texttt{forward} method implements the inverse transformation $\cX \rightarrow \cZ$ and the likelihood contribution $\cV(\vx,\vz)$, since this is what is needed during the forward pass of backpropagation used for training.

In Sec. \ref{subsec:vae} we show an implementation of a VAE as a stochastic transformation, while in Sec. \ref{subsec:dequantization} and Sec. \ref{subsec:augmentation} we show implementations of dequantization and variational data augmentation as surjective transformations. Finally, in Sec. \ref{subsec:example}, we show an example of how to construct an augmented normalizing flow through composition of SurVAE layers.

\subsection{VAE}
\label{subsec:vae}

We implement VAEs as a composable stochastic transformation.

\begin{lstlisting}
class VAE(StochasticTransform):
    '''A variational autoencoder layer.'''

    def __init__(self, decoder, encoder):
        super(VAE, self).__init__()
        self.decoder = decoder
        self.encoder = encoder

    def forward(self, x):
        z, log_qz = self.encoder.sample_with_log_prob(context=x)
        log_px = self.decoder.log_prob(x, context=z)
        ldj = log_px - log_qz
        return z, ldj

    def inverse(self, z):
        x = self.decoder.sample(context=z)
        return x
\end{lstlisting}

\newpage
\subsection{Dequantization}
\label{subsec:dequantization}

We implement \texttt{UniformDequantization}, which may be used to convert between discrete and continuous variables, as a generative rounding surjection.

\begin{lstlisting}
class UniformDequantization(Surjection):
    '''A uniform dequantization layer.'''

    def forward(self, x):
        z = x.float() + torch.rand_like(x)
        ldj = torch.zeros(x.shape[0])
        return z, ldj

    def inverse(self, z):
        x = z.floor().long()
        return x
\end{lstlisting}

\subsection{Augmentation}
\label{subsec:augmentation}

We implement \texttt{Augment}, a generative tensor slicing surjection, which may be used to construct e.g. augmented normalizing flows \citep{huang2020, chen2020}.

\begin{lstlisting}
class Augment(Surjection):
    '''An augmentation layer.'''

    def __init__(self, encoder, split_size):
        super(Augment, self).__init__()
        self.encoder = encoder
        self.split_size = split_size

    def forward(self, x):
        z2, log_qz2 = self.encoder.sample_with_log_prob(context=x)
        z = torch.cat([x, z2], dim=1)
        ldj = -log_qz2
        return z, ldj

    def inverse(self, z):
        x, z2 = torch.split(z, self.split_size, dim=1)
        return x
\end{lstlisting}

\subsection{Example: Augmented Normalizing Flows}
\label{subsec:example}

We showcase here the simplicity of implementing an augmented normalizing flow using the SurVAE flow framework. In Listing \ref{lst:flow}, a simple normalizing flow consisting of 2 coupling layers is constructed. In Listing \ref{lst:aflow}, this is extended by adding an \texttt{Augment} surjection, resulting in an augmented flow. 

\begin{minipage}{.45\textwidth}
\raggedright
\begin{lstlisting}[caption=A basic flow., label=lst:flow]
Flow(base_dist=Normal((2,)),
     transforms=[
     
        CouplingBijection(),
        Reverse(),
        CouplingBijection(),
     ])
\end{lstlisting}
\end{minipage}
\begin{minipage}{.55\textwidth}
\raggedleft
\begin{lstlisting}[caption=An augmented flow., label=lst:aflow]
Flow(base_dist=Normal((4,)),
     transforms=[
        Augment(Normal((2,)), (2,2)),
        CouplingBijection(),
        Reverse(),
        CouplingBijection(),
     ])
\end{lstlisting}
\end{minipage}

     

\begin{figure}[ht]
    \centering

\begin{subfigure}[b]{0.31\textwidth}
    \centering
    \caption*{\textbf{Data}}
    \vspace{-2mm}
    \includegraphics[width=\textwidth]{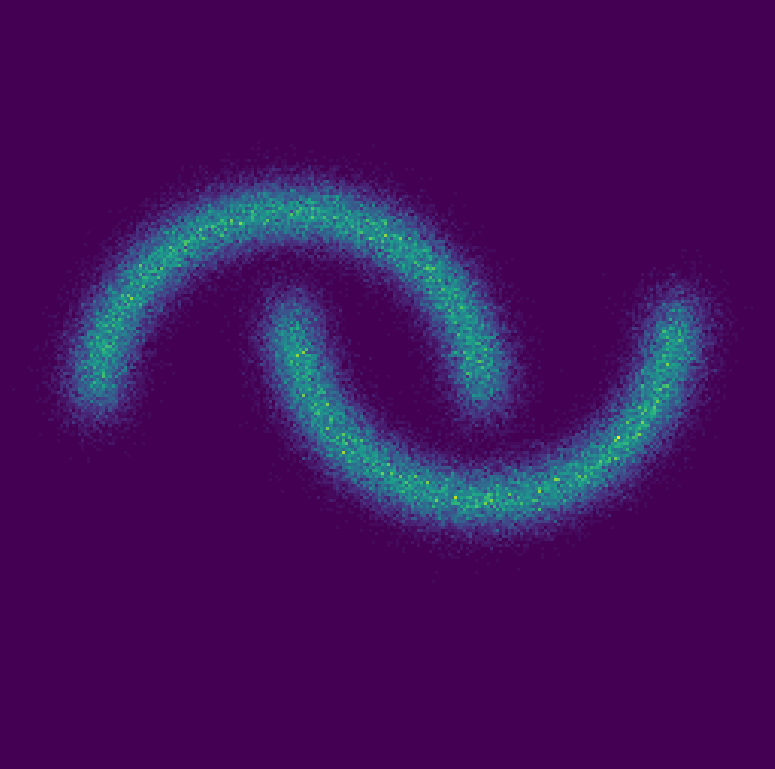}
\end{subfigure}
\begin{subfigure}[b]{0.31\textwidth}
    \centering
    \caption*{\textbf{Flow}}
    \vspace{-2mm}
    \includegraphics[width=\textwidth]{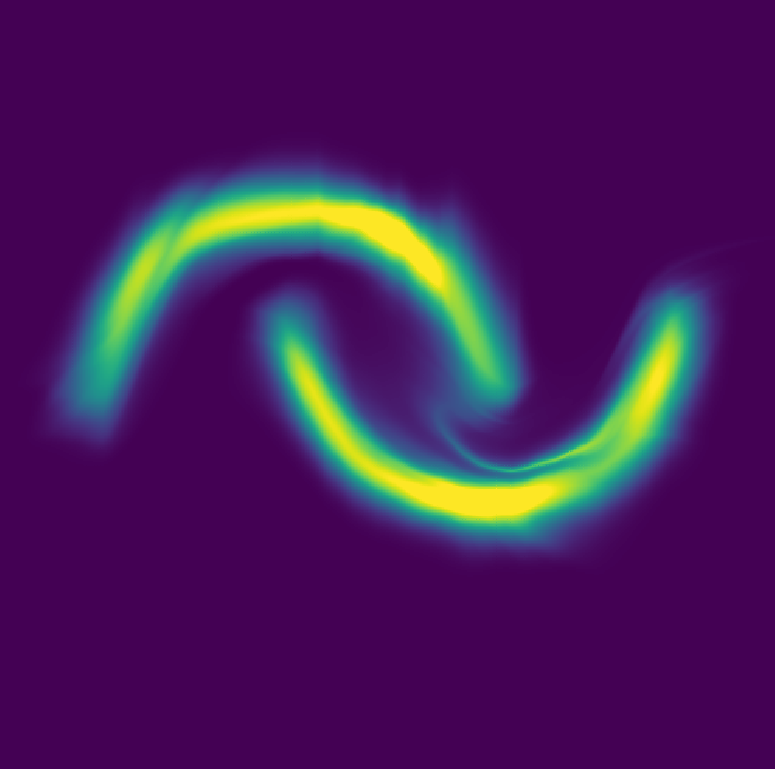}
\end{subfigure}
\begin{subfigure}[b]{0.31\textwidth}
    \centering
    \caption*{\textbf{Augmented Flow}}
    \vspace{-2mm}
    \includegraphics[width=\textwidth]{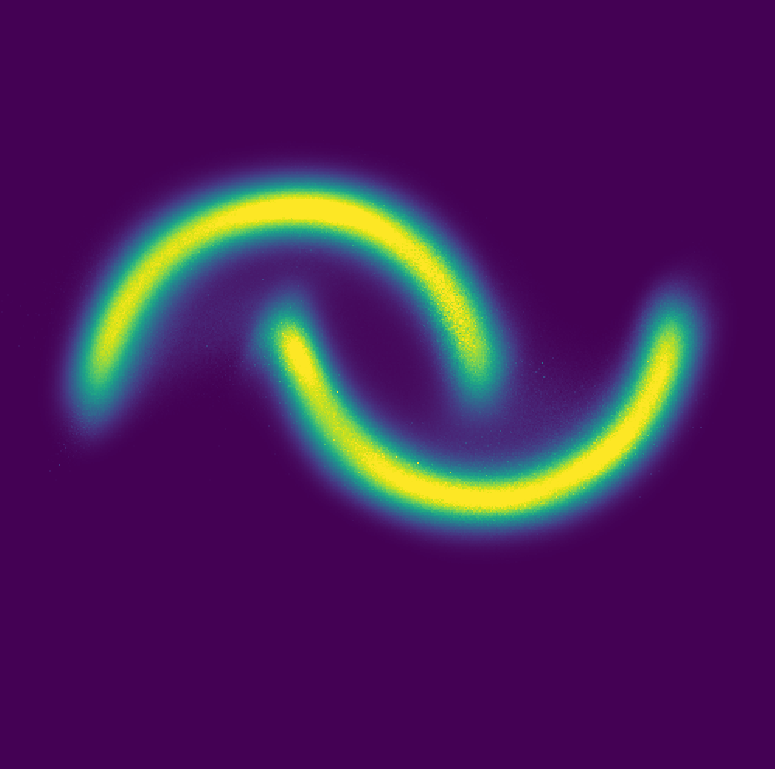}
\end{subfigure}

\begin{subfigure}[b]{0.31\textwidth}
    \centering
    \includegraphics[width=\textwidth]{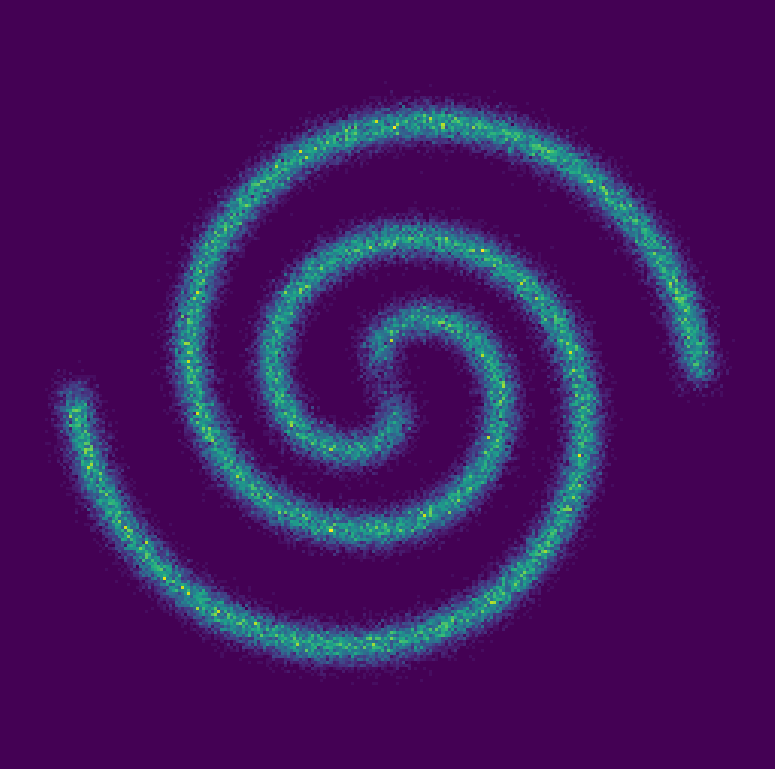}
\end{subfigure}
\begin{subfigure}[b]{0.31\textwidth}
    \centering
    \includegraphics[width=\textwidth]{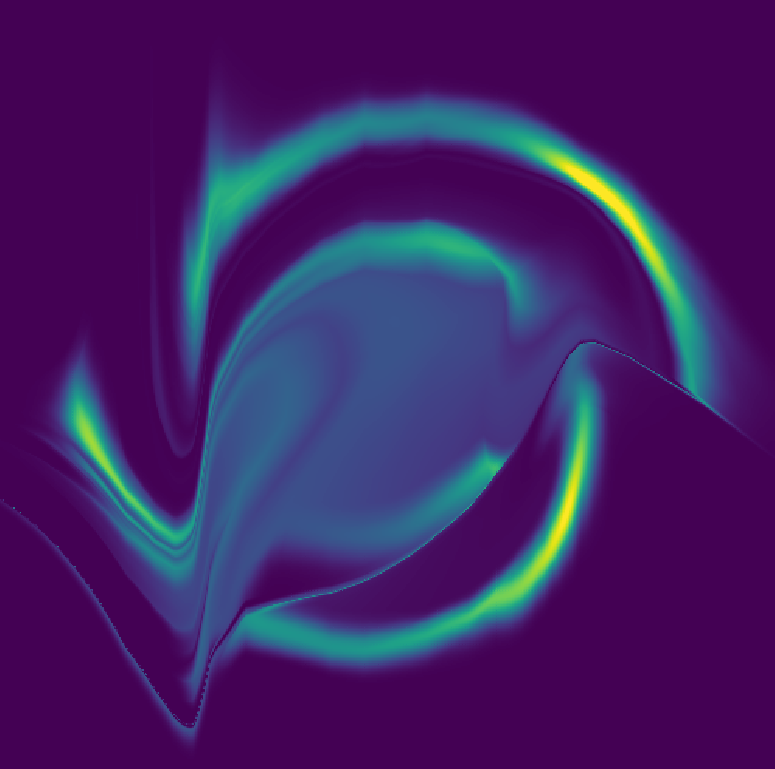}
\end{subfigure}
\begin{subfigure}[b]{0.31\textwidth}
    \centering
    \includegraphics[width=\textwidth]{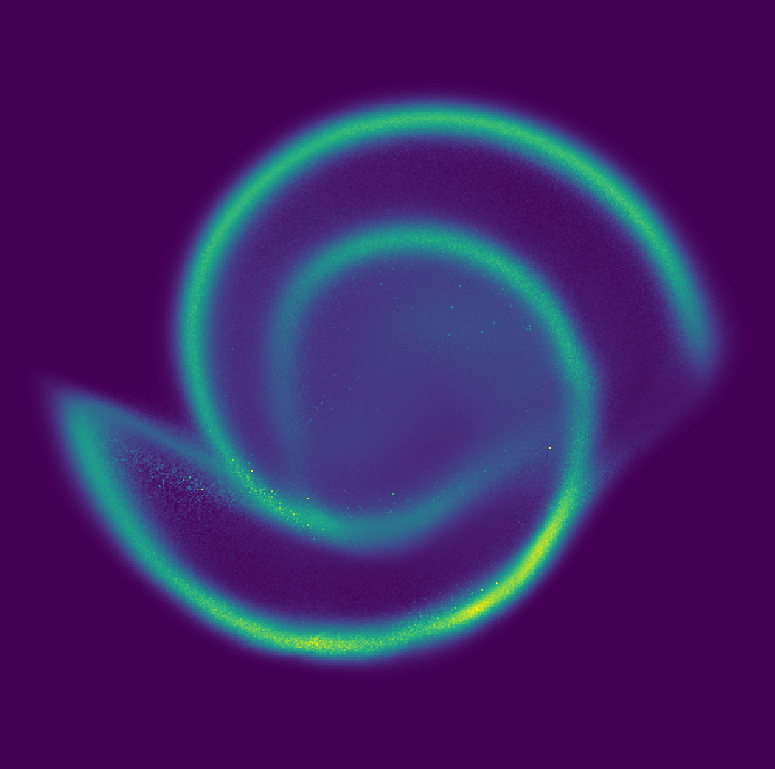}
\end{subfigure}

\begin{subfigure}[b]{0.31\textwidth}
    \centering
    \includegraphics[width=\textwidth]{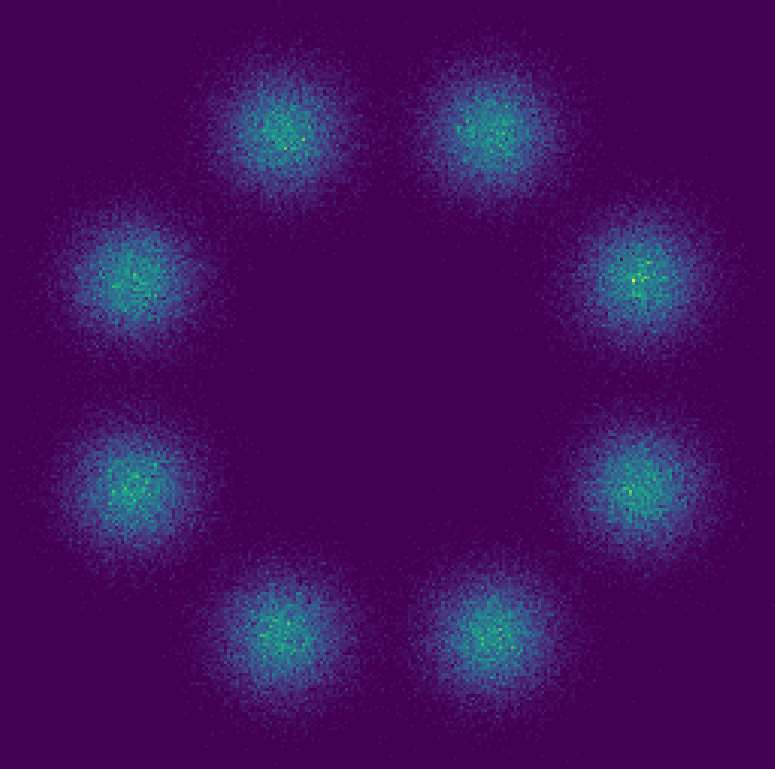}
\end{subfigure}
\begin{subfigure}[b]{0.31\textwidth}
    \centering
    \includegraphics[width=\textwidth]{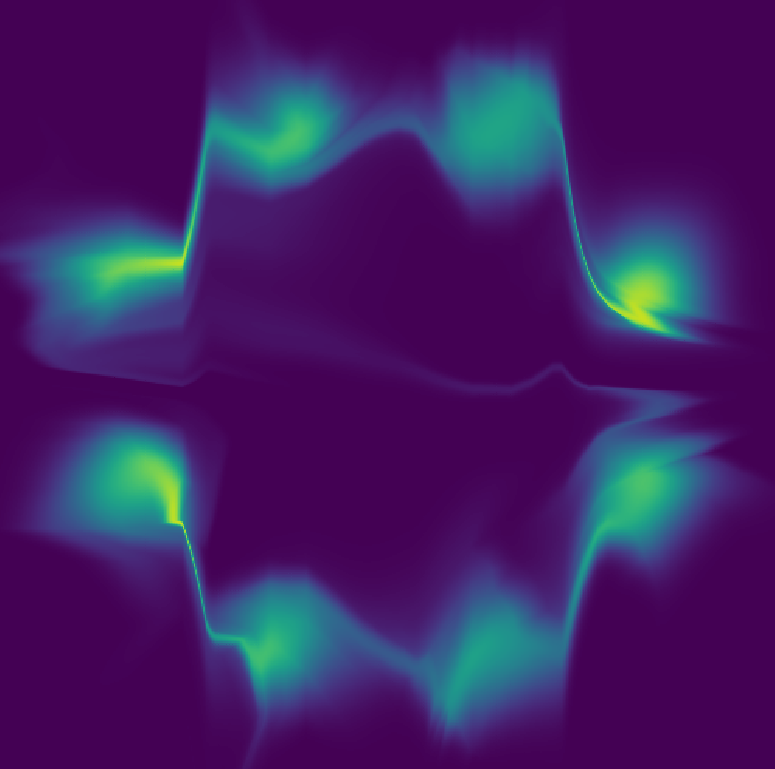}
\end{subfigure}
\begin{subfigure}[b]{0.31\textwidth}
    \centering
    \includegraphics[width=\textwidth]{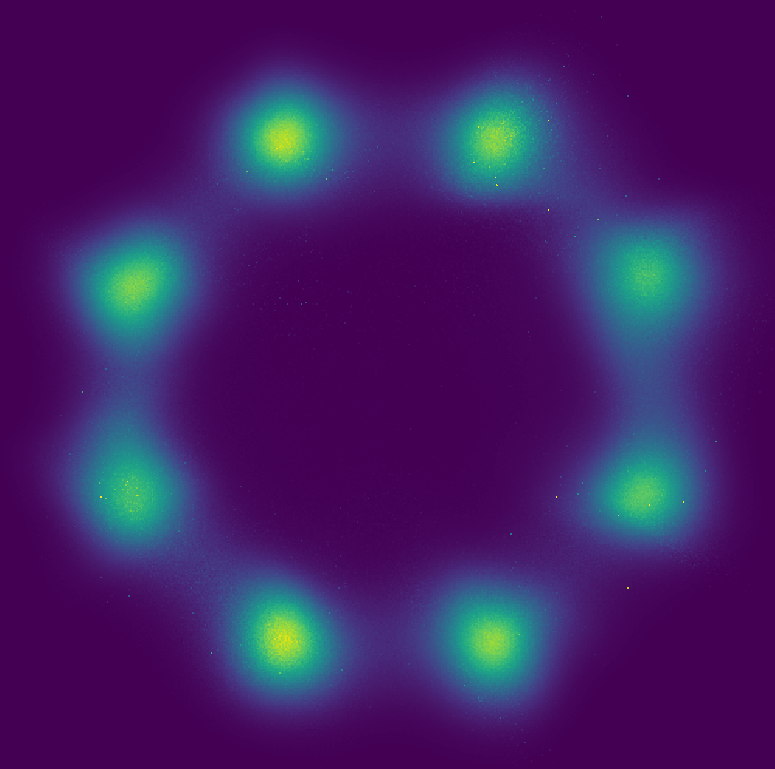}
\end{subfigure}

\begin{subfigure}[b]{0.31\textwidth}
    \centering
    \includegraphics[width=\textwidth]{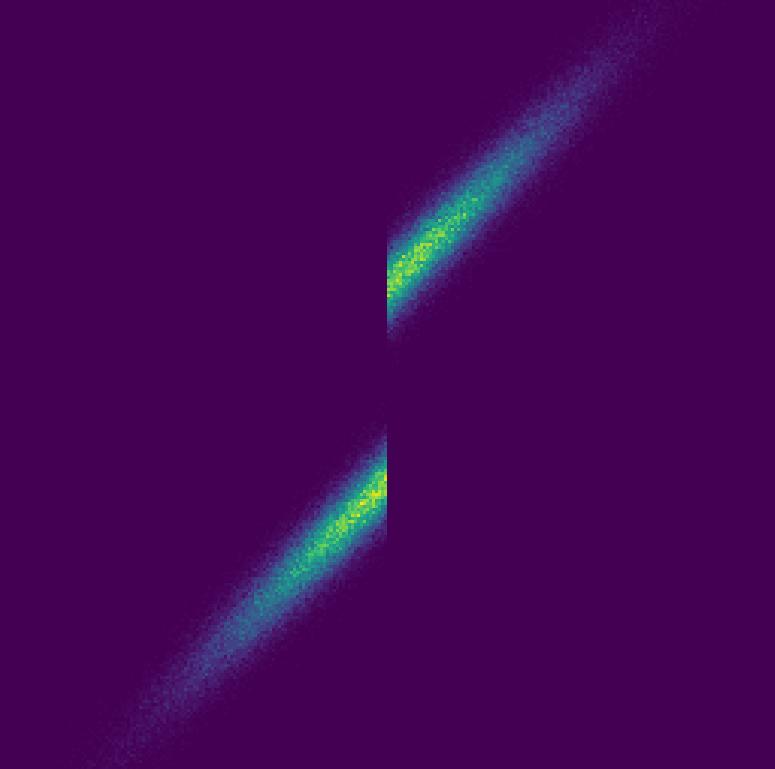}
\end{subfigure}
\begin{subfigure}[b]{0.31\textwidth}
    \centering
    \includegraphics[width=\textwidth]{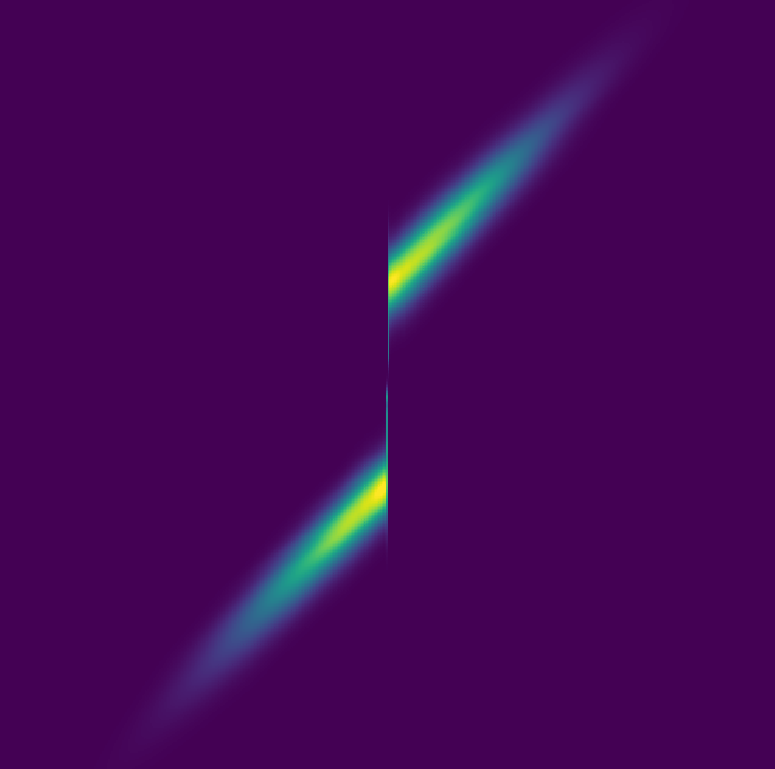}
\end{subfigure}
\begin{subfigure}[b]{0.31\textwidth}
    \centering
    \includegraphics[width=\textwidth]{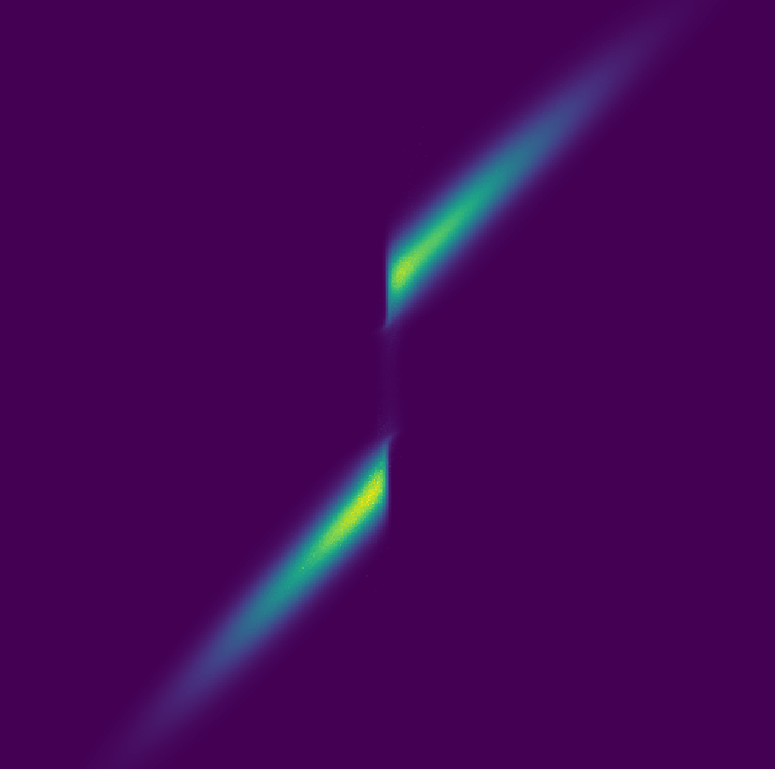}
\end{subfigure}

    \caption{Augmented flows show improved capabilities over flows at modelling 2D densities, especially where there are disconnected components. With SurVAE flows, augmented flows are implemented by adding a single surjective augmentation layer to the flow as shown in Listing \ref{lst:aflow}.}
    \label{fig:augflow}
\end{figure}

Using the models in Listing \ref{lst:flow} and Listing \ref{lst:aflow}, we compare a standard coupling flow with a simple extension using an additional \texttt{Augment} layer. We use 4 coupling layers instead of 2 and train models both using identical setups 10000 iterations each. Augmented flows have improved capabilites of modelling data with disconnected components. In Fig. \ref{fig:augflow}, we observe that the augmented flows tend to place their mass more out in a more "clean" fashion and thus demonstrate improved ability to model complicated 2D densities.

\newpage
\null\newpage
\section{Experimental Details}
\label{app:experiments}

We here give more details on the experiments. 
For further details, see our open-source code\footnote{\url{https://github.com/didriknielsen/survae_flows}}.

\subsection{Synthetic Data}
\label{app:synthetic}

\textbf{Data.} We used 4 synthetic datasets, \texttt{checkerboard}, \texttt{corners}, \texttt{gaussians} and \texttt{circles}. For each syntheric dataset, 128000 samples were used as a training set and 128000 more samples as a test set. The \texttt{checkerboard} dataset is anti-symmetric, while the 3 others are symmetric.

\textbf{Training.} We used the Adam optimizer \citep{kingma2014} with a learning rate of $10^{-3}$. All models were trained for 10000 iterations (10 epochs) using a batch size of 128.

\textbf{Baseline.} The baseline flow is a composition of 4 affine coupling bijections with the ordering reversed in-between. The coupling layers are parameterized by MLPs with hidden units (200,100) and ReLU activations. The base distribution is a standard Gaussian.

\textbf{Symmetric AbsFlow.} For the symmetric datasets, AbsFlow uses all the same layers as the baseline. In addition, an \texttt{abs} surjection is added, followed by and inverse softplus (\texttt{gaussians}) or logit (\texttt{checkerboard} and \texttt{corners}). In the generative direction, the \texttt{abs} surjection randomly samples the sign with equal probabilities. The extra layers contain no parameters, and the AbsFlow thus have the exact same number of parameters as the baseline. 

\textbf{Anti-Symmetric AbsFlow.} For the anti-symmetric dataset, AbsFlow uses only a single \texttt{abs} surjection and a uniform base distribution. In the generative direction, a classifier network learns the probabilities of sampling the sign conditioned on $\vz$. This classifier network is, like the coupling layer networks, an MLP with (200,100) hidden units and ReLU activations. In this case, the AbsFlow thus has \char`\~ $1/4$ the number of parameters.

\subsection{Point Cloud Data}
\label{app:point_cloud}

\textbf{Data.} We used the \texttt{SpatialMNIST} dataset \citep{edwards2017}. This dataset was constructed by, for each digit in the \texttt{MNIST} dataset, sampling 50 points according to the normalized pixel intensities. We used the official code\footnote{\url{https://github.com/conormdurkan/neural-statistician}} to construct the dataset. We split the dataset into parts of 50000-10000-10000 for training, validation and test (without shuffling). Each data example is a set of 50 2D points which we represent as a tensor of shape \texttt{(2,50)}.

\textbf{Training.} Both models were trained for 500 epochs using a batch size of 128. We used the Adam optimizer \citep{kingma2014} with an initial learning rate of $10^{-3}$. The learning rate was warmed up linearly for 2000 iterations and the decayed by 0.995 every epoch. All models were trained using a single GPU for about 40 hours.

\textbf{Evaluation.} SortFlow allows exact computation of the likelihood, while the PermuteFlow only allows computation of lower bounds. We evaluated PermuteFlow using the IWBO (importance weighted bound) \citep{burda2016} using $k=1000$ importance samples. PermuteFlow obtains an ELBO of -5.32 PPLL and an IWBO of -5.30 PPLL, while SortFlow obtains an exact log-likelihood of -5.53 PPLL.

\textbf{Hyperparameters.} We tuned the dropout rate using the validation set. We considered dropout rates of $\{0.0, 0.05, 0.1, 0.2, 0.3\}$ for both models. We found 0.1 to work best for PermuteFlow, while 0.2 worked best for SortFlow.

\textbf{PermuteFlow.} We used a flow of an initial stochastic permutation layer followed by 32 steps with ActNorm layers \citep{kingma2018} in-between.
Each step consisted of 1) an affine coupling bijection which transforms a the first half tensor \texttt{(1,50)} conditioned on the other half \texttt{(1,50)}, 2) reversing the order along the spatial dimension, 3) an affine coupling bijection which transforms the first half tensor \texttt{(2,25)} conditioned on the other half \texttt{(2,25)}, 4) a stochastic permutation along the point dimension. Each of the coupling bijections are parametersized by Transformer networks \citep{vaswani2017} \emph{without} positional encoding. The Transformers used 2 blocks, with $d_{\mathrm{model}} = 64$, $d_{\mathrm{ff}} = 256$ and 8 attention heads.

\textbf{SortFlow.} This follows the setup of PermuteFlow, with the following changes: 1) The initial stochastic permutation is replaced by a sorting layer. 2) The stochastic permutations in the flow are swapped with fixed permutations (sampled at random once, before training). 3) The Transformers make use of a learned positional encoding, since the sorting layer enforces a canonical ordering of the points.

\subsection{Image Data}
\label{app:image}

\textbf{Data.} We used the \texttt{CIFAR-10}, \texttt{ImageNet $32 \times 32$} and \texttt{ImageNet $64 \times 64$} datasets. The \texttt{CIFAR-10} dataset comes pre-split in 50000 training examples and 10000 test examples. The \texttt{ImageNet} datasets also come pre-split in 1,281,149 training examples and 49,999 validation examples. We use these splits and report results for the test set of \texttt{CIFAR-10} and the validations sets of \texttt{ImageNet $32 \times 32$} and \texttt{ImageNet $64 \times 64$}.

\textbf{Training.} We used the \texttt{Adamax} optimizer \citep{kingma2014} with an initial learning rate of $10^{-3}$ and a batch size of 32. The learning rate was linearly warmed up for 5000 iterations. For \texttt{CIFAR-10}, the models were first trained for 500 epochs with the learning rate decayed by 0.995 every epoch. Next, the models were "cooled down" for an additional 50 epochs with a smaller learning rate of $2 \cdot 10^{-5}$. For the \texttt{ImageNet} datasets, the models were first trained for 25 epochs (\texttt{ImageNet $32 \times 32$}) and 20 epochs (\texttt{ImageNet $64 \times 64$}) with the learning rate decayed by 0.95 every epoch. Next, the models were "cooled down" for an additional 2 epochs with a smaller learning rate of $5 \cdot 10^{-5}$.
The \texttt{CIFAR-10} and \texttt{ImageNet $32 \times 32$} models were trained on a single GPU for about 2 weeks, while the \texttt{ImageNet $64 \times 64$} models were trained using 4 GPUs for about 3 weeks. We provide pre-trained model checkpoints in our open-source code. Note that data augmentation was applied during training of the \texttt{CIFAR-10} models, including random flipping and rotations. See code for more details. 

\textbf{Evaluation.} The \texttt{CIFAR-10} models were evaluated using the IWBO (importance weighted bound) \citep{burda2016} using $k=1000$ importance samples. The \texttt{ImageNet} models were evaluated using the ELBO (which corresponds to the IWBO with $k=1$ importance sample). 

\textbf{Baseline.} For \texttt{CIFAR-10} and \texttt{ImageNet $32 \times 32$}, the flow uses 2 scales with 12 steps/scale. For \texttt{ImageNet $64 \times 64$}, the flow uses 3 scales with 8 steps/scale. Each step consists of an affine coupling bijection \citep{dinh2017} and an invertible $1 \times 1$ convolution \citep{kingma2018}. All models are trained with variational dequantization \citep{ho2019} and an initial squeezing layer \citep{dinh2017} to increase the number of channels from 3 to 12. The coupling bijections are parameterized by DenseNets \citep{huang2017}.

\textbf{MaxPoolFlow.} The MaxPoolFlow uses the exact same setup as the baseline, but replaces the tensor slicing surjection with a max pooling surjection. In the generative direction, we used the simplest possible choice: Each input pixel is equally likely to be copied to any of the pixels in its corresponding $2 \times 2$ patch. The remaining 3 elements are sampled such that the copied value remains the largest: They are set equal to this maximum value minus noise from a standard half-normal distribution (i.e. Gaussian distribution with only positive values). We used simple choices containing \emph{no extra parameters} in order to facilitate more fair comparison. Note that the max pooling layer could be potentially be improved by using more sophisticated choices for the distribution for sampling the remaining elements, $p(\vx_{-k}|\vz)$, and/or by using a classifier, $p(\vk|\vz)$, to predict the indices $\vk$.

\newpage
\section{Additional Samples}
\label{app:samples}

Samples from SurVAE flows trained on \texttt{CIFAR-10}, \texttt{ImageNet $32 \times 32$} and \texttt{ImageNet $64 \times 64$} using either max pooling or tensor slicing for downsampling are shown in Fig. \ref{fig:c10_samples}, Fig. \ref{fig:i32_samples} and Fig. \ref{fig:i64_samples}, respectively.

\newcommand{\sampleswidth}{0.4\textwidth}
\begin{figure}[!ht]
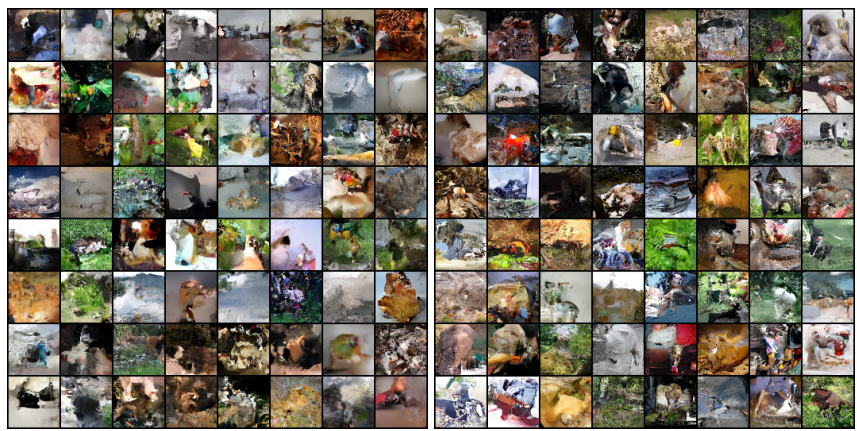

    \centering
    \begin{subfigure}[b]{\sampleswidth}
            \centering
            \includegraphics[width=\textwidth]{figures/samples/c10_max.png}
    \label{fig:c10_max}
    \subcaption{Max Pooling.}
    \end{subfigure}
    \begin{subfigure}[b]{\sampleswidth}
            \centering
            \includegraphics[width=\textwidth]{figures/samples/c10_non.png}
    \label{fig:c10_non}
    \subcaption{No Pooling.}
    \end{subfigure}

    \caption{Unconditional samples from SurVAE flows trained on \texttt{CIFAR-10}.}
    \label{fig:c10_samples}
\end{figure}

\begin{figure}[!ht]
    \centering
    \begin{subfigure}[b]{\sampleswidth}
            \centering
            \includegraphics[width=\textwidth]{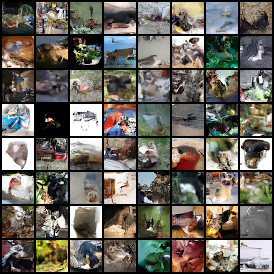}
    \label{fig:i32_max}
    \subcaption{Max Pooling.}
    \end{subfigure}
        \begin{subfigure}[b]{\sampleswidth}
            \centering
            \includegraphics[width=\textwidth]{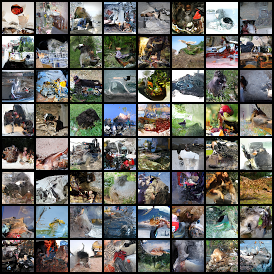}
    \label{fig:i32_non}
    \subcaption{No Pooling.}
    \end{subfigure}

    \caption{Unconditional samples from SurVAE flows trained on \texttt{ImageNet $32 \times 32$}.}
    \label{fig:i32_samples}
\end{figure}

\begin{figure}[!ht]
    \centering
    \begin{subfigure}[b]{\sampleswidth}
            \centering
            \includegraphics[width=\textwidth]{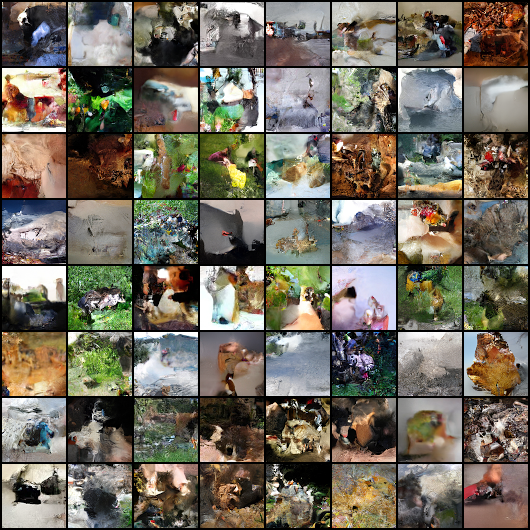}
    \label{fig:i64_max}
    \subcaption{Max Pooling.}
    \end{subfigure}
        \begin{subfigure}[b]{\sampleswidth}
            \centering
            \includegraphics[width=\textwidth]{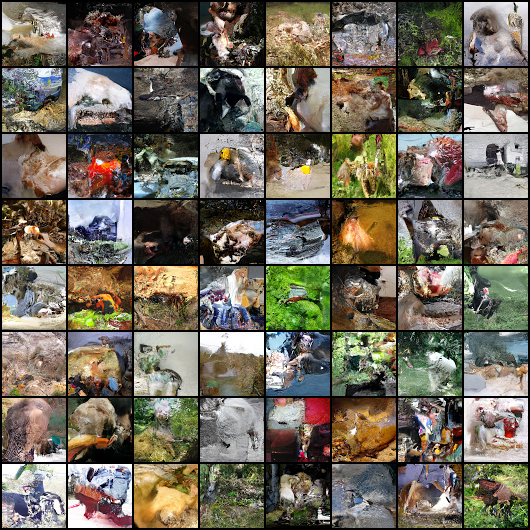}
    \label{fig:i64_non}
    \subcaption{No Pooling.}
    \end{subfigure}

    \caption{Unconditional samples from SurVAE flows trained on \texttt{ImageNet $64 \times 64$}.}
    \label{fig:i64_samples}
\end{figure}

\end{document}